\documentclass[runningheads]{llncs}
\usepackage{graphicx}
\usepackage{amsmath,amssymb} 
\usepackage{color}\usepackage[width=122mm,left=12mm,paperwidth=146mm,height=193mm,top=12mm,paperheight=217mm]{geometry}

\usepackage{times}
\usepackage{epsfig}
\usepackage{subfigure}
\usepackage{float}
\usepackage{bm}
\usepackage{xspace}
\usepackage{hyperref}
\usepackage{slashbox}
\usepackage{array}

\makeatletter
\DeclareRobustCommand\onedot{\futurelet\@let@token\@onedot}
\def\@onedot{\ifx\@let@token.\else.\null\fi\xspace}

\def\eg{\emph{e.g}\onedot} 
\def\ie{\emph{i.e}\onedot}

\def\etal{\emph{et al}\onedot}
\makeatother

\begin{document}
\pagestyle{headings}
\mainmatter
\title{Affine Subspace Representation for Feature Description} 

\author{Zhenhua Wang, Bin Fan and Fuchao Wu}
\institute{National Laboratory of Pattern Recognition, Institute of Automation\\
Chinese Academy of Sciences, 100190, Beijing, China \\
{\tt\small \{wzh,bfan,fcwu\}@nlpr.ia.ac.cn}}

\maketitle

\begin{abstract}
This paper proposes a novel Affine Subspace Representation (ASR) descriptor to deal with affine distortions induced by viewpoint changes.
Unlike the traditional local descriptors such as SIFT, ASR inherently encodes local information of multi-view patches, making it robust to affine distortions while maintaining a high discriminative ability.
To this end, PCA is used to represent affine-warped patches as PCA-patch vectors for its compactness and efficiency.
Then according to the subspace assumption, which implies that the PCA-patch vectors of various affine-warped patches
of the same keypoint can be represented by a low-dimensional linear subspace,
the ASR descriptor is obtained by using a simple subspace-to-point mapping. Such a linear subspace representation could accurately capture
the underlying information of a keypoint (local structure) under multiple views without sacrificing its distinctiveness.
To accelerate the computation of ASR descriptor, a fast approximate algorithm is proposed by moving the most computational part (\ie, warp patch under various affine transformations) to an offline training stage. Experimental results show that ASR is not only better than the state-of-the-art descriptors under various image transformations, but also performs well without a dedicated affine invariant detector when dealing with viewpoint changes.
\end{abstract}

\section{Introduction}

Establishing visual correspondences is a core problem in computer vision.
A common approach is to detect keypoints in different images and
construct keypoints' local descriptors for matching.
The challenge lies
in representing keypoints with discriminative descriptors, which are also
invariant to photometric and geometric transformations.

Numerous methods have been proposed in the literature to tackle such problems in a certain degree.
The scale invariance is often achieved by estimating the characteristic scales of keypoints.
The pioneer work is done by Lindeberg~\cite{LINDEBERG:1998}, who proposes a systematic methodology for automatic scale selection
by detecting the keypoints in multi-scale representations.
Local extremas over scales of different combinations of $\gamma$-normalized
derivatives indicate the presence of characteristic local structures.
Lowe~\cite{LOWE:2004} extends the idea of Lindeberg by
selecting scale invariant keypoints in Difference-of-Gaussian (DoG) scale space.
Other alternatives are SURF~\cite{BAY:2006}, BRISK~\cite{Leutenegger2011},
Harris-Laplacian and Hessian-Laplacian~\cite{Mikolajczyk04}.
Since these methods are not designed for affine invariance,
their performances drop quickly under significant viewpoint changes.
To deal with the distortion induced by viewpoint changes,
some researchers propose to detect regions covariant to the affine transformations.
Popular methods include Harris-Affine~\cite{Mikolajczyk04}, Hessian-Affine~\cite{MIKOLAJCZYK:2005},
MSER~\cite{MATAS:2002},
EBR and IBR~\cite{Tuytelaars04}.
They estimate the shapes of elliptical regions and
normalize the local neighborhoods into circular regions
to achieve affine invariance.
Since the estimation of elliptical regions are not accurate,
ASIFT~\cite{morel2009} proposes to simulate all image views under the full affine space
and match the SIFT features extracted in all these simulated views to establish correspondences.
It improves the matching performance at the cost of a huge computational complexity.

\begin{figure}[htb]
\centering
\includegraphics[width=0.65\textwidth]{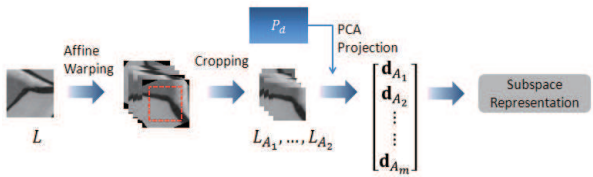}
\caption{The workflow of constructing ASR descriptor.}
\label{subfig:asb-naive}
\end{figure}

This paper aims to tackle the affine distortion by developing a novel Affine Subspace Representation (ASR) descriptor, which effectively models the inherent information of a local patch among multi-views. Thus it can be combined with any detector to match images with viewpoint changes, while traditional methods usually rely on an affine-invariant detector, such as Harris-Affine + SIFT.
Rather than estimating the local affine transformation, the main innovation of this paper lies in directly building descriptor by exploring
the local patch information under multiple views.
Firstly, PCA (Principle Component Analysis) is applied to all the warped patches of a keypoint under various viewpoints to obtain a set of patch representations. These representations are referred to as \emph{PCA-patch vectors} in this paper.
Secondly, each set of PCA-patch vectors is represented by a low-dimensional linear subspace
under the assumption that PCA-patch vectors computed from various affine-warped patches of the same keypoint span a linear subspace.
Finally, the proposed Affine Subspace Representation (ASR) descriptor is obtained by using a subspace-to-point mapping. Such a linear subspace representation could efficiently capture the underlying local information of a keypoint under multiple views,
making it capable of dealing with affine distortions.
The workflow our method is summarized in Fig.~\ref{subfig:asb-naive}, each step of which will be elaborated in Section 3.
To speedup the computation, a fast approximate algorithm is proposed by
removing most of its computational cost to an offline learning stage (the details will be introduced in Section 3.3). This is the second contribution of this paper. Experimental evaluations on image matching with various transformations have demonstrated that the proposed descriptors can achieve state-of-the-art performance. Moreover, when dealing with images with viewpoint changes, ASR performs rather well without a dedicated affine detector, validating the effectiveness of the proposed method.

The rest of this paper is organized as follows:
Section~\ref{sec:related_work} gives an overview of related works.
The construction of the proposed ASR descriptor as well as its fast computation algorithm are elaborated in Section~\ref{sec:our_approach}. Some details in implementation is given in Section 4.
Experimental evaluations are reported in Section~\ref{sec:experiments} and finally we conclude the paper
in Section~\ref{sec:conclusion}.

\section{Related Work\label{sec:related_work}}

Lindeberg and Garding~\cite{lindeberg1997shape} presented a methodology for reducing affine shape distortion.
The suggested approach is to adapt the shape of smoothing kernel to the local image structure
by measuring the second moment matrix.
They also developep a method for extracting blob-like affine features
with an iterative estimation of local structures.
Based on the work of Lindeberg,
Baumberg~\cite{baumberg2000} adapted the local shapes of keypoints at fixed scales and locations
, while Mikolajczyk and Schmid~\cite{Mikolajczyk04} iteratively
estimated the affine shape as well as the location and scale.
Tuytelaars and Van Gool~\cite{Tuytelaars04} proposed two affine invariant detectors.
The geometry-based method detects Harris corners and extracts edges close to such keypoints.
Several functions are then chosen to determine a parallelogram
spanned by the nearby two edges of the keypoint.
The intensity-based method extracts local extremas in intensity as anchor points.
An intensity function along rays emanating from these anchor points is used
to select points where this function reaches an extremum.
All these selected points are linked to enclose an affine covariant region which is further replaced
by an ellipse having the same shape moments up to the second moments.
Matas \etal~\cite{MATAS:2002} developed an efficient affine invariant detector
based on the concept of extremal regions.
The proposed maximally stable extremal regions are produced by a watershed algorithm and their
boundaries are used to fit elliptical regions.

Since the accuracy of affine shape estimation is not guaranteed,
Morel and Yu~\cite{morel2009} presented a new framework for affine invariant image matching named ASIFT.
They simulated all possible affine distortion
caused by the change of camera optical axis orientation from a frontal position, and extract SIFT features
on all these simulated views. The SIFT features on all simulated views are matched to find correspondences.
Since ASIFT has to compute
SIFT on lots of simulated views and make use of an exhaustive search on all possible views,
it suffers a huge computational complexity. Although a similar view simulation method of ASIFT is used in our method, here it is for a totally different purpose: warping local patch of a keypoint under multiple views to extract PCA-patch vectors for keypoint description. Therefore, our method does not suffer from the huge computational burden as in ASIFT.
Hintersoisser \etal~\cite{Hinterstoisser2011} proposed two learning based methods to
deal with full perspective transformation.
The first method trains a Fern classifier~\cite{ozuysal2010fast}
with patches seen under different viewing conditions in order
to deal with perspective variations, while the second one
uses a simple nearest neighbors classifier on a set of
\lq\lq mean patches\rq\rq
 that encodes the average of the keypoints appearance over a limited set of poses.
However, an important limitation of these two methods is
that they can not scale well with the size of keypoints database.
Moreover, they both need a fronto-parallel view for
training and the camera internal parameters for computing the camera pose relative to the keypoint.

The most related work to this paper is SLS~\cite{Hassner2012}, which describes each pixel as a set of SIFT descriptors extracted at multiple scales.
Our work extends SLS to deal with affine variations. Moreover, we propose to use PCA-patch vector as a compact intermediate representation of the warped patch instead of SIFT. The main advantages are two-folds: (a) fast, because PCA-patch vector is fast to compute while computing SIFT is much slower; (b) since computing PCA vector is a linear operation, it leads to the proposed fast algorithm.

\section{Our Approach\label{sec:our_approach}}

\subsection{Multiple View Computation \label{ss:multiple view_computation}}

As the projective transformation induced by camera motion around a smooth surface
can be locally approximated by an affine transformation,
we locally model the apparent deformations arising from the camera motions by affine transformations.
In order to deal with affine distortions, we propose to integrate local patch
information under various affine transformations for feature description
rather than estimating the local affine transformation (\eg,~\cite{Tuytelaars04,Mikolajczyk04}).

Since we employ scale-invariant detector to select keypoints,
we first extract a local patch at the given scale around each keypoint
and then resize it to a uniform size of $s_l\times s_l$.
To deal with linear illumination changes,
the local patch is usually normalized to have zero mean and unit variance.
Here we skip this step since the subsequent computation of linear subspace
is invariant to linear illumination changes.

The local patch is aligned by the local dominant orientation to achieve invariance to in-plane rotation.
In order to efficiently estimate such orientations,
we sample some pattern points in the local patch similar to BRISK~\cite{Leutenegger2011}.
The dominant orientation is then estimated by the average gradient direction of
all the sampling points:
\begin{equation}
\overline {\bf{g}}  = (
{1/n_p\sum\limits_{i = 1}^{{n_p}} {{g_x}({\bf{p_i}})} }, {1/n_p\sum\limits_{i = 1}^{{n_p}} {{g_y}({\bf{p_i}})} }
),
\end{equation}

where $n_p$ is the number of sampling points, $\overline {\bf{g}}$ is the average gradients,
$g_x(\bf{p_i})$ and $g_y(\bf{p_i})$ are the $x$-directional and $y$-directional gradients
of $i^{th}$ sampling point $\bf{p_i}$ respectively.
Since there are only a few sample points, \eg, $n_p=60$ in our experiments,
the orientation can be estimated very fast.

Let $L$ be the aligned reference patch around a keypoint at a given scale,
the warped patch under an affine transformation $A$ is computed by:
\begin{equation}
{L_A} = w(L,A),
\label{eq:LA_orig}
\end{equation}
where $w(\cdot,A)$ is the warping function using transformation $A$.
To avoid the case that some parts of the warped patch may not visible in the reference patch,
we take the reference patch a little larger in practice.
Hence, Eq. (\ref{eq:LA_orig}) can be re-written as:
\begin{equation}
{L_A} = p(w(L,A)),
\label{eq:LA_rewritten}
\end{equation}
where $p(\cdot)$ is a function
that extracts a small central region from the input matrix.

To encode the local information of each $L_A$,
we propose to use a simple PCA based representation for its compactness and efficiency.
By using PCA, the local patch is projected into the eigenspace and
the largest $n_d$ principal component coordinates are taken to represent the patch, \ie, the PCA-patch vector.
Mathematically, the PCA-patch vector ${\bf{d_A}}$ for $L_A$ can be computed as :

\begin{equation}
\begin{split}
{\bf{d_A}} =  {P_d}^T {vec(L_A)} =  f(L_A)
\end{split}
\label{eq:PCA_patch_des},
\end{equation}
where
$P_d$ is the learned PCA projection matrix,
$vec(\cdot)$ denotes vectorization of a matrix,
and $f(\cdot)={P_d}^T vec(\cdot)$.
By substituting Eq. (\ref{eq:LA_rewritten}),
Eq. (\ref{eq:PCA_patch_des}) can be rewritten as:
\begin{equation}
{\bf{d}_A} = f(p(w(L,A))).
\label{eq:DA}
\end{equation}

The idea of using PCA for feature description is not novel,
\eg, PCA-SIFT descriptor in~\cite{YAN2004} and GLOH descriptor in~\cite{Mikolajczyk:2005:descriptor}.
Here we only use such a technique to
effectively generate a set of compact vectors as the intermediate representations.
Further representation of the keypoint will be explored based on these intermediate representations.

\subsection{Subspace Representation\label{ss:subspace}}
Suppose there are $m$ parameterized affine transformations
to warp a local patch, we can get a PCA-patch vector set $\mathcal{D}=\{{\bf{d}_{A_m}}\}$
for a keypoint by the above approach.
Inspired by Hassner \etal.\cite{Hassner2012} who dealt with
scale invariant matching by using a linear subspace representation
of SIFT descriptors extracted on multiple scales,
we proposed to construct a subspace model to
represent the PCA-patch vectors extracted on multiple views.

\begin{figure}[htb]
\begin{minipage}[t]{0.48\textwidth}
\centering
\includegraphics[width=0.48\textwidth]{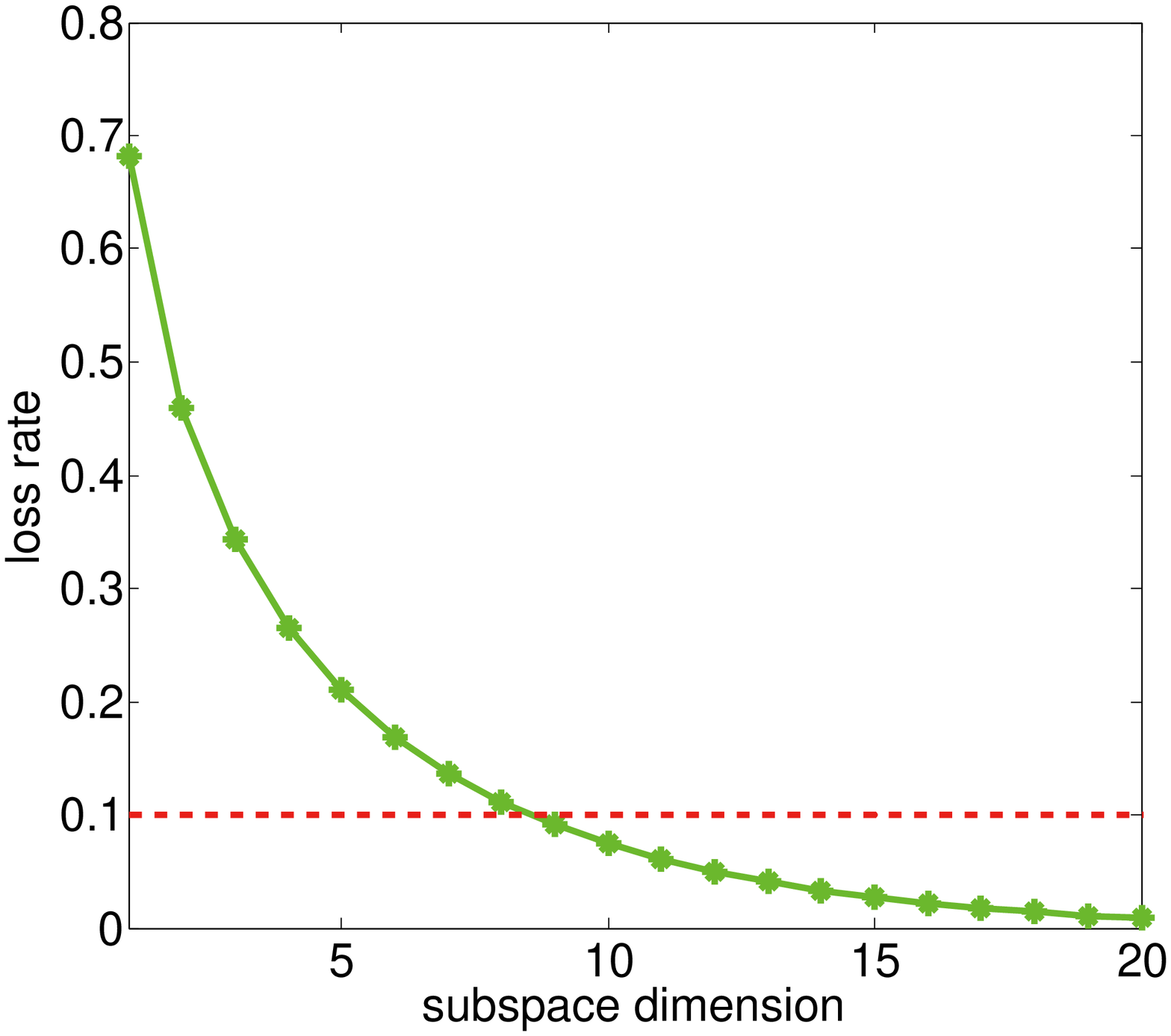}
\caption{Averaged loss rate as a function of the subspace dimension.
The patch size is $21\times21$ and the dimension of PCA-patch vector is $24$.\label{fig:subspace_dim}}
\end{minipage}
\hspace{2ex}
\begin{minipage}[t]{0.48\textwidth}
\centering
\includegraphics[width=0.48\textwidth]{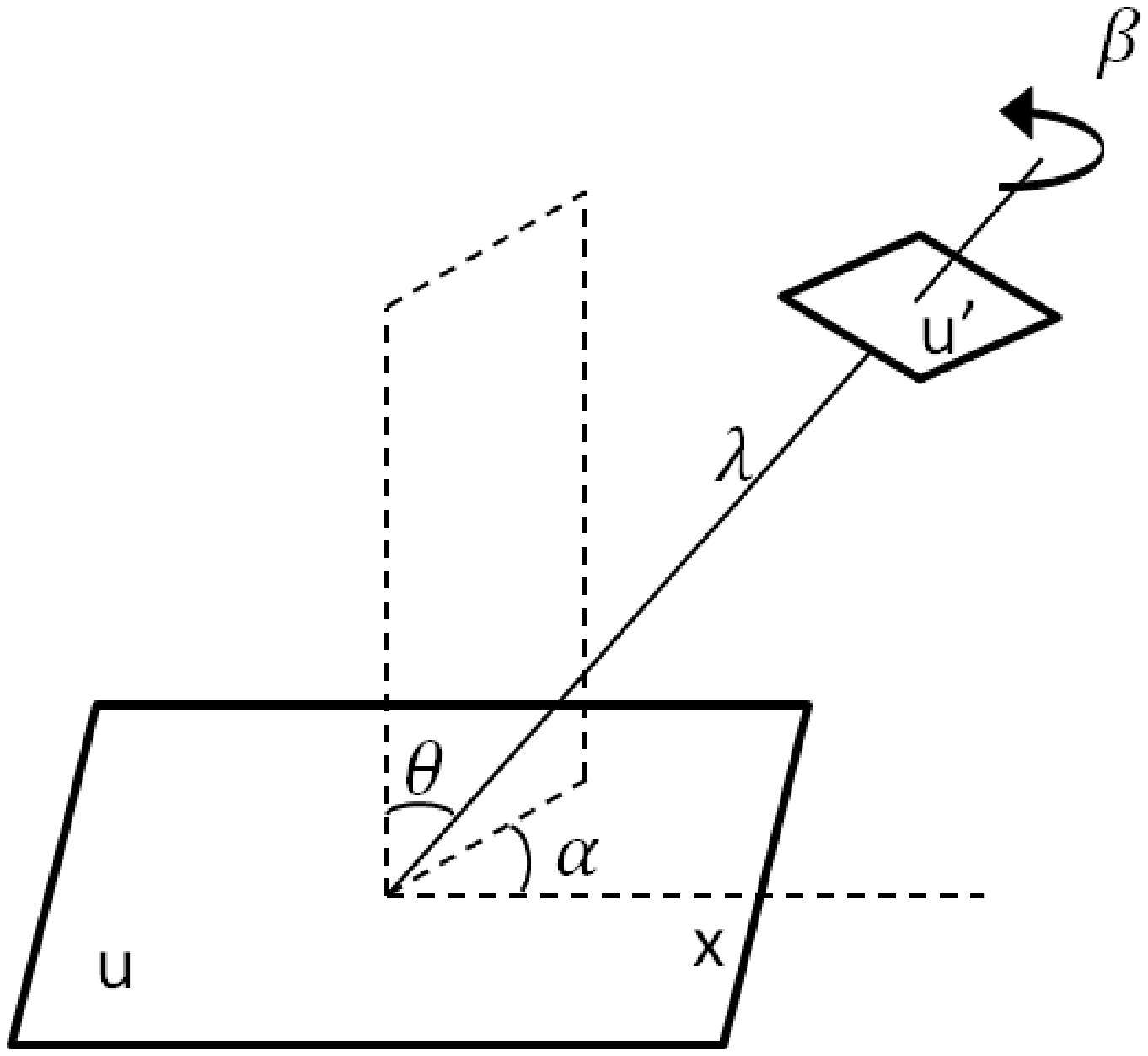}
\caption{
Geometric interpretation of the decomposition in Eq. (\ref{eq:affine_model}). See text for details.
\label{fig:affine_model}}
\end{minipage}
\end{figure}

The key observation is that the PCA-patch vectors
extracted under various affine transformations of a same keypoint
approximately lie on a low-dimensional linear subspace.
To show this point,
we conducted statistical analysis on the reconstruction loss rates\footnote{It is defined as the rate between reconstruction error and the original data, while the reconstruction error is the squared distance between the original data and its reconstruction by PCA.}
of PCA-patch vectors for about 20,000 keypoints detected from images randomly downloaded from the Internet.
For each keypoint, its PCA-patch vector set is computed and used to estimate a subspace by PCA.
Then the reconstruction loss rates of each set by using different numbers of subspace basis are recorded.
Finally, the loss rates of all PCA-patch vector sets are averaged.
Fig.~\ref{fig:subspace_dim} shows how the averaged loss rate is changed with different subspace dimensions.
It can be observed that a subspace of 8 dimensions is enough to approximate the 24 dimensional PCA-patch vector set
with 90\% information kept in average.
Therefore, we choose to use a $n_s$-dimensional linear subspace to represent $\mathcal{D}$. Mathematically,
{\begin{equation}
[{{\bf{d}}_{{A_1}}}, \cdots ,{{\bf{d}}_{{A_m}}}] \approx
[{\widehat {\bf{d}}_1}, \cdots ,{\widehat {\bf{d}}_{{n_s}}}]
\left[ \begin{array}{*{20}{c}}
{{b_{11}}, \cdots ,{b_{1m}}}\\
{\begin{array}{*{20}{c}}
 \vdots & \ddots & \vdots
\end{array}}\\
{{b_{{n_s}1}}, \cdots ,{b_{{n_s}m}}}
\end{array}\right] ,
\end{equation}}
where ${\widehat {\bf{d}}_1}, \cdots ,{\widehat {\bf{d}}_{{n_s}}}$
are basis vectors spanning the subspace and
$b_{ij}$ are the coordinates in the subspace.
By simulating enough affine transformations,
the basis ${\widehat {\bf{d}}_1}, \cdots ,{\widehat {\bf{d}}_{{n_s}}}$
can be estimated by PCA.

Let $\mathcal{D}_k$ and $\mathcal{D}_{k'}$ be the PCA-patch vector sets of
keypoints $k$ and $k'$ respectively, the distance between
$\mathcal{D}_k$ and $\mathcal{D}_{k'}$ can be measured by the distance between
corresponding subspaces $\mathbb{D}_k$ and $\mathbb{D}_{k'}$.
As shown in~\cite{edelman1998},
all the common distances between two subspaces are defined based on the principal angles.
In our approach, we use the Projection Frobenius Norm defined as:
\begin{equation}
dist(\mathbb{D}_k,\mathbb{D}_{k'})
= \left\| {\sin {\bm{\psi }}} \right\|_2
= \frac{1}{\sqrt 2} {\left\| {{\widehat{D}_k}{\widehat{D}_k}^T -
 {\widehat{D}_{k'}}{\widehat{D}_{k'}}^T} \right\|_F} ,
\end{equation}
where ${\sin {\bm{\psi }}}$ is a vector of sines of the principal angles between subspaces
$\mathbb{D}_k$ and $\mathbb{D}_{k'}$, $\widehat{D}_k$ and $\widehat{D}_{k'}$ are
matrixes whose columns are basis vectors of
 subspaces $\mathbb{D}_k$ and $\mathbb{D}_{k'}$ respectively.

To obtain a descriptor representation of the subspace,
similar to~\cite{Hassner2012} we employ the subspace-to-point mapping proposed by Basri \etal~\cite{basri2011}.
Let $\widehat D$ be the matrix composed of orthogonal basis of subspace $\mathbb{D}$,
the proposed ASR descriptor can be obtained by mapping the projection matrix
$Q=\widehat{D}\widehat{D}^T$ into vectors.
Since $Q$ is symmetric, the mapping $h(Q)$ can be defined as
rearranging the entries of $Q$ into a vector by taking the upper triangular
portion of $Q$, with the diagonal entries scaled by $1\slash\sqrt 2$.
Mathematically, the ASR descriptor {\bf{q}} is
\begin{equation}
{\bf{q}} = h(Q) = (\frac{{{q_{11}}}}{{\sqrt 2 }},{q_{12}}, \cdots ,{q_{1{n_d}}},
\frac{{{q_{22}}}}{{\sqrt 2 }},{q_{23}}, \cdots ,
\frac{{{q_{{n_d}{n_d}}}}}{{\sqrt 2 }}) ,
\end{equation}
where $q_{ij}$ are elements of $Q$, and $n_d$ is the dimension of the PCA-patch vector.
Thus the dimension of $\bf{q}$ is $n_d*(n_d+1)/2$.

By such mapping, it is worth noting that the Projection Frobenius Norm distance between subspaces $\mathbb{D}_k$ and $\mathbb{D}_{k'}$
is equal to the Euclidean distance between the corresponding ASR descriptors ${{\bf{q}}_k}$
and ${{\bf{q}}_{k'}}$:
\begin{equation}
dist(\mathbb{D}_k,\mathbb{D}_{k'}) = {\left\| {{{\bf{q}}_k} - {{\bf{q}}_{k'}}} \right\|_2}.
\end{equation}

It is worth noting that ASR is inherently invariant to linear illumination changes.
Suppose $\mathcal{D}=\{{\bf{d}_{A_m}}\}$ is the set of PCA-patch vectors for a keypoint,
while $\mathcal{D}'=\{{\bf{d}_{A_m}}'\}$ is its corresponding set after linear illumination changes.
For each element in $\mathcal{D}$, ${\bf{d}_{A_m}} = a\times{\bf{d}_{A_m}}'+b$ where $a$ and $b$ parameterize the linear illumination changes.
Let $cov(\mathcal{D})$ and $cov(\mathcal{D}')$ be their covariant matrixes,
it is easy to verify that $cov(\mathcal{D}) = a^2\times cov(\mathcal{D}')$. Therefore, they have the same eigenvectors.
Since it is the eigenvectors used for ASR construction, the obtained ASR for $\mathcal{D}$ and $\mathcal{D}'$ will be identical.

\subsection{Fast Computation}

Due to the high computational burden of warping patches,
it would be very inefficient to compute a set of PCA-patch vectors extracted under various affine transformations
by utilizing Eq. (\ref{eq:DA}) directly.

In~\cite{Hinterstoisser2011}, Hinterstoisser \etal proposed a
method to speed up the computation of warped patches under different camera poses
based on the linearity of warping function.
We found that their method could be easily extended to speed up the computation of any linear descriptor of the warped patches.
According to this observation,
we develop a fast computation method of ${{\bf{d}}_A}$ at the cost of a little accuracy degradation in this section.

Similar to~\cite{Hinterstoisser2011},
we firstly approximate $L$ by its principal components as:
\begin{equation}
L \approx \overline{L}  + \sum\limits_{i = 1}^{{n_l}} {a_i{L_i}}
\label{eq:LA_approx},
\end{equation}
where $n_l$ is the number of principal components,
$L_i$ and $a_i$ are the principal components and the projection coordinates respectively.

Then, by substituting Eq. (\ref{eq:LA_approx}) into Eq. (\ref{eq:DA}) , it yields:
\begin{equation}
{{\bf{d}}_A} \approx f(p(w(\overline{L}  + \sum\limits_{i = 1}^{{n_l}} {{a_i}{L_i}} ,A))).
\label{eq:DA_approx}
\end{equation}

\begin{figure}[htb]
\centering
\includegraphics[width=0.70\textwidth]{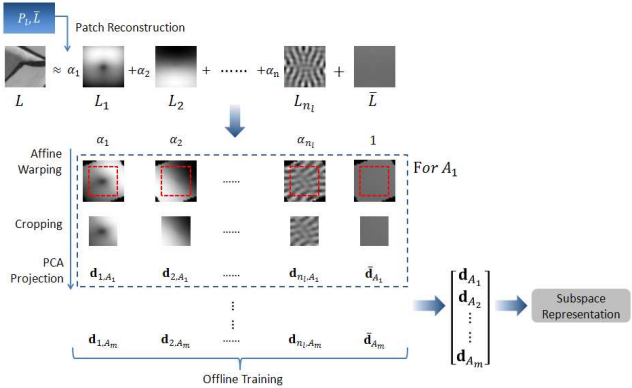}
\caption{Fast computation strategy for constructing ASR descriptor.}
\label{subfig:asb-fast}
\end{figure}

Note that the warping function $w(\cdot,A)$ is essentially a permutation of the pixel intensities
between the reference patch and the warped patch. It implies that $w(\cdot,A)$ is actually a linear transformation.
Since $p(\cdot)$ and $f(\cdot)$ are also linear functions, Eq. (\ref{eq:DA_approx}) can be re-written as:

\begin{equation}
{{\bf{d}}_A} \approx f(p(w(\overline{L} ,A))) + \sum\limits_{i = 1}^{{n_l}} {{a_i}f(p(w({L_i},A)})) = {\overline {\bf{d}} _A} + \sum\limits_{i = 1}^{{n_l}} {{a_i}{{\bf{d}}_{i,A}}},
\label{eq:DA_final}
\end{equation}
where
\begin{equation}
  \begin{split}
   {\overline {\bf{d}} _A} = &\ f(p(w(\overline{L} ,A))) \\
    {{\bf{d}}_{i,A}} = &\ f(p(w({L_i},A)))
   \end{split}.
   \label{eq:DA_final2}
\end{equation}
Fig.~\ref{subfig:asb-fast} illustrates the workflow of such a fast approximated algorithm.

Although the computation of ${\overline {\bf{d}} _A}$ and ${{\bf{d}}_{i,A}}$
is still time consuming, it can be previously done in an offline learning stage.
At run time,
we simply compute the projection coordinates ${\bf{a}} = {({a_1}, \cdots ,{a_{{n_l}}})^T}$ of
the reference patch $L$ by
\begin{equation}
{\bf{a}} = {P_l}^T vec(L) ,
\end{equation}
where $P_l$ is the learned projection matrix consisting of $L_i$.
Then, ${\bf{d}_A}$ can be computed by
a linear combination of ${{\bf{d}}_{i,A}}$ and ${\overline {\bf{d}} _A}$.

Obviously, this approach combines the patch warping and representation into one step,
and moves most of the computational cost to the offline learning stage.
Compared to the naive way in Eq. (\ref{eq:DA}),
it significantly reduces the running time.
We refer to ASR descriptor computed by such a fast approximate algorithm as ASR-fast descriptor,
while the original one is referred to as ASR-naive descriptor.

\section{Notes on Implementation}
\subsection{Parameterization of Affine Transformation\label{ss:affine_model}}

As shown in~\cite{morel2009}, any 2D affine transformation $A$ with strictly positive determinant
which is not a similarity has a unique decomposition:

\begin{equation}
A = \lambda R(\alpha )T(t)R(\beta ) = \lambda \left[\begin{array}{*{20}{c}}
{\cos \alpha }&{ - \sin \alpha }\\
{\sin \alpha }&{\cos \alpha }
\end{array}\right]\left[\begin{array}{*{20}{c}}
t&0\\
0&1
\end{array}\right]\left[\begin{array}{*{20}{c}}
{\cos \beta }&{ - \sin \beta }\\
{\sin \beta }&{\cos \beta }
\end{array}\right],
\label{eq:affine_model}
\end{equation}

where $\lambda >0$, $R$ is a rotation matrix, $\alpha \in[0,\pi)$, $\beta \in[0,2\pi)$,
$T$ is a diagonal matrix with $t>1$.

Fig.~\ref{fig:affine_model} gives a geometric interpretation of this decomposition:
$u$ is the object plane, $u'$ is the image plane,
$\alpha$ (longitude) and $\theta=\arccos 1\slash t$ (latitude) are the camera viewpoint angles,
$\beta$ is the camera in-plane rotation, and $\lambda$ is the zoom parameter.
The projective transformation from image plane $u'$ to object plane $u$ can be
approximated by the affine transformation in Eq. (\ref{eq:affine_model}).

Since the scale parameter $\lambda$ can be estimated by scale-invariant detectors
and the in-plane rotation $\beta$ can be aligned by local dominant orientation,
we only sample the longitude angle $\alpha$ and the tilt $t$.

For $\alpha$, the sampling range is $[0,\pi)$ as indicated by the decomposition in Eq. (\ref{eq:affine_model}).
The sampling step $\Delta \alpha = \alpha_{k+1}-\alpha_k$ is determined by considering the overlap
between the corresponding ellipses of adjacent samplings.
More specifically, for an affine transformation $A_{t,\alpha}$ with tilt $t$ and longitude $\alpha$,
the corresponding ellipse is $e_{t,\alpha}=A^{T}_{t,\alpha}A_{t,\alpha}$.
Let $\varepsilon (e_{t,\alpha}, e_{t,\alpha+\Delta \alpha})$ denotes the overlap rate
between $e_{t,\alpha}$ and $e_{t,\alpha+\Delta \alpha}$,
it can be proved that $\varepsilon (e_{t,\alpha}, e_{t,\alpha+\Delta \alpha})$ is
a decreasing function of $\Delta \alpha$ when $t>1 \land \Delta \alpha \in [0,\pi\slash 2)$.
We can choose the sampling step $\Delta \alpha$ as the max value that satisfies
$\varepsilon (e_{t,\alpha}, e_{t,\alpha+\Delta \alpha}) > T_o$ where $T_o$
is a threshold that controls the minimal overlap rate required for the corresponding ellipses of adjacent samplings.
The larger $T_o$ is, the more $\alpha$ will be sampled.

For $t$, the sampling range is set to $[1,4]$ to make the latitude angle $\theta=\arccos 1\slash t$
range from $0^\circ$ to $75^\circ$.
Thus, the sampling step $\Delta t=t_{k+1}\slash t_{k}$ is $4^{\frac{1}{n_t-1}}$
where $n_t$ is the sampling number of $t$.

Setting these sampling values is not a delicate matter. To show this point, we have investigated the influence of different sampling strategies for $\alpha$ and $t$ on image pair of \emph{'trees 1-2'} of the Oxford dataset~\cite{Oxford}.
Fig.~\ref{subfig:affine model t} shows the performance of ASR-naive by varying $n_t$ ($3$, $5$, $7$ and $9$) when $T_o=0.8$.
It can be seen that $n_t=5$, $n_t=7$ and $n_t=9$ are comparable and they are better than $n_t=3$.
Therefore, we choose $n_t=5$ since it leads to the least number of affine transformations.
Under the choice of $n_t=5$, we also test its performance on various $T_o$ ($0.6$, $0.7$, $0.8$ and $0.9$) and the result
is shown in Fig.~\ref{subfig:affine model alpha}.
Although $T_o=0.9$ performs the best, we choose $T_o=0.8$ to make a compromise between accuracy and sparsity (complexity).
According to the above sampling strategy, we totally have $44$ simulated affine transformations.
Note that the performance is robust to these values in a wide range. Similar observations can be obtained in other test image pairs.

\subsection{Offline Training}
From Section 3, it can be found there are three cases in which PCA is utilized:
\begin{itemize}
\item[(1)]PCA is used for raw image patch representation to obtain a PCA-patch vector for each affine-warped image patches.
\item[(2)]PCA is used to find subspace basis of a set of PCA-patch vectors for constructing ASR descriptor.
\item[(3)]PCA is used to find principal components to approximate a local image patch $L$ for fast computation (c.f. Eq. (\ref{eq:LA_approx}).
\end{itemize}
In cases of (1) and (3), several linear projections is required. More specifically, $n_d$ principal projections are used for PCA-patch vector computation and $n_l$ principal components are used to approximate a local image patch. These PCA projections are learned in an offline training stage. In this stage, the PCA projection matrix $P_d$ in Eq. (\ref{eq:PCA_patch_des}),
${{\bf{d}}_{i,A}}$ and ${\overline {\bf{d}} _A}$ in Eq. (\ref{eq:DA_final2}) are computed by using about 2$M$ patches detected on $17125$ training images provided by PASCAL VOC 2012. Thus the training images are significantly different from those used for performance evaluation in Section~\ref{sec:experiments}.

\section{Experiments\label{sec:experiments}}
In this section, we conduct experiments to show the effectiveness of the proposed method.
Firstly, we study the potential impact of different parameter settings on the performance of the proposed method.
Then, we test on the widely used Oxford dataset~\cite{Oxford} to show its superiority to the state-of-the-art local descriptors. With the image pairs under viewpoint changes in this dataset, we also demonstrate that it is capable of dealing with affine distortion without an affine invariant detector, and is better than the traditional method, e.g., building SIFT descriptor on Harris Affine region. To further show its performance in dealing with affine distortion, we conduct experiments on a larger dataset (Caltech 3D Object Dataset~\cite{Moreels:2007}), containing a large amount of images of different 3D objects captured from different viewpoints. The detailed results are reported in the following subsections.

\begin{figure}[htb]
\begin{minipage}[t]{0.48\textwidth}
\centering
\subfigure[varying $n_t$ when $T_o=0.8$]
{\includegraphics[width=0.48\textwidth]{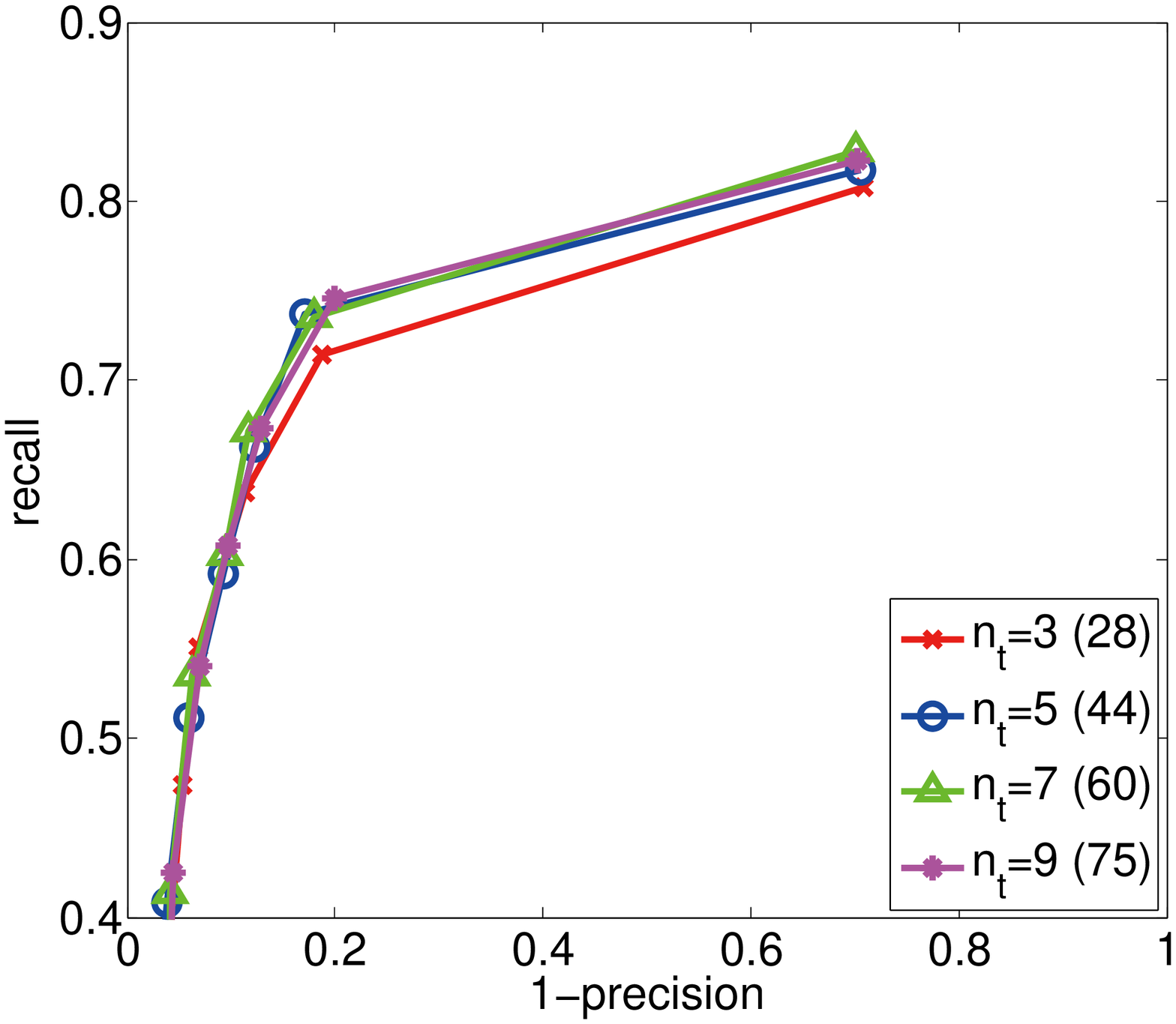}\label{subfig:affine model t}}
\subfigure[varying $T_o$ when $n_t=5$]
{\includegraphics[width=0.48\textwidth]{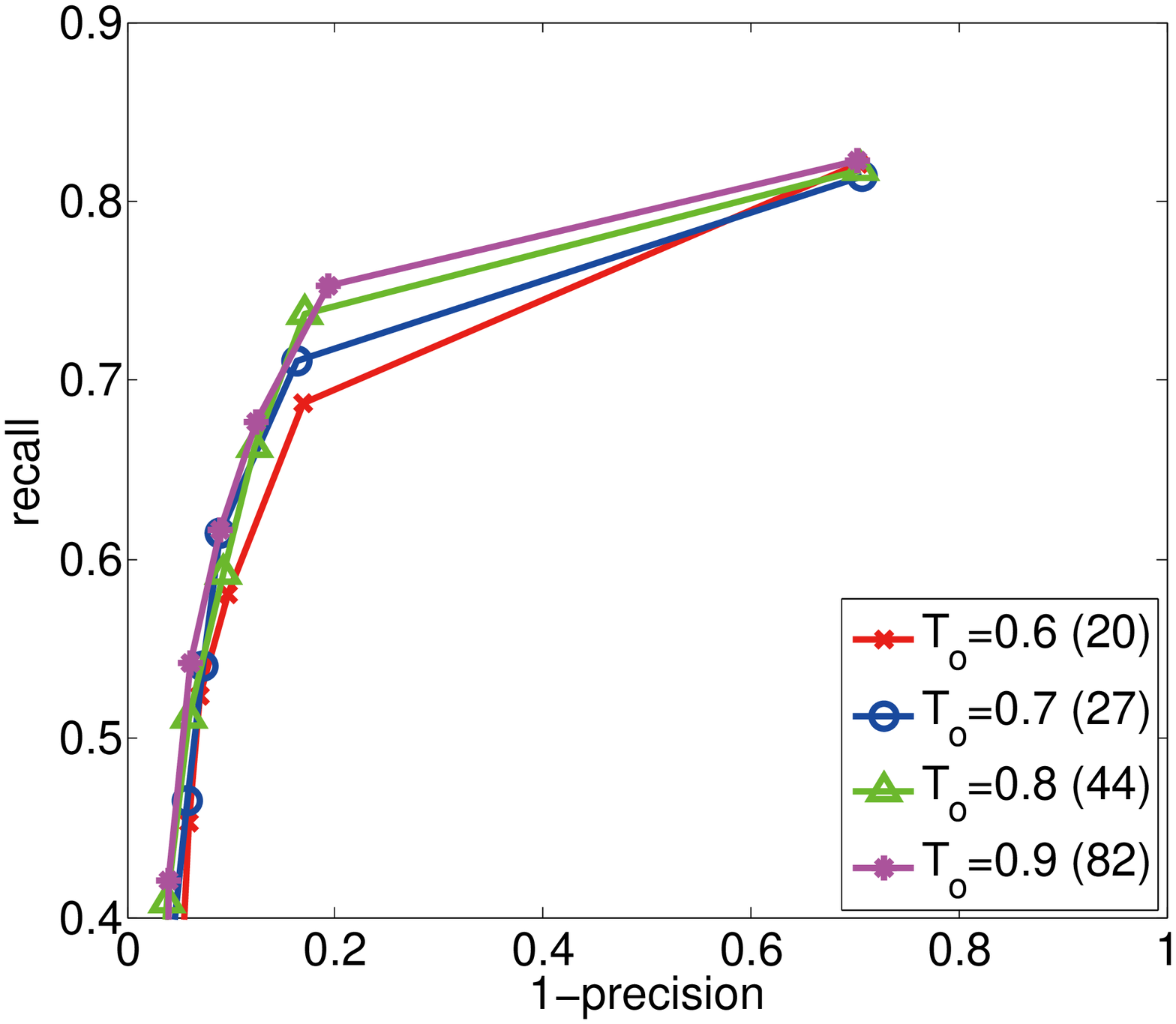}\label{subfig:affine model alpha}}
\caption{Performance comparison of ASR descriptor on DoG keypoints under different sampling strategies.
 The number of simulated affine transformations is enclosed in the parenthesis.\label{fig:parameters_1}}
\end{minipage}
\hspace{2ex}
\begin{minipage}[t]{0.48\textwidth}
\centering
\subfigure[varying $n_d$ and $n_s$]
{\includegraphics[width=0.48\textwidth]{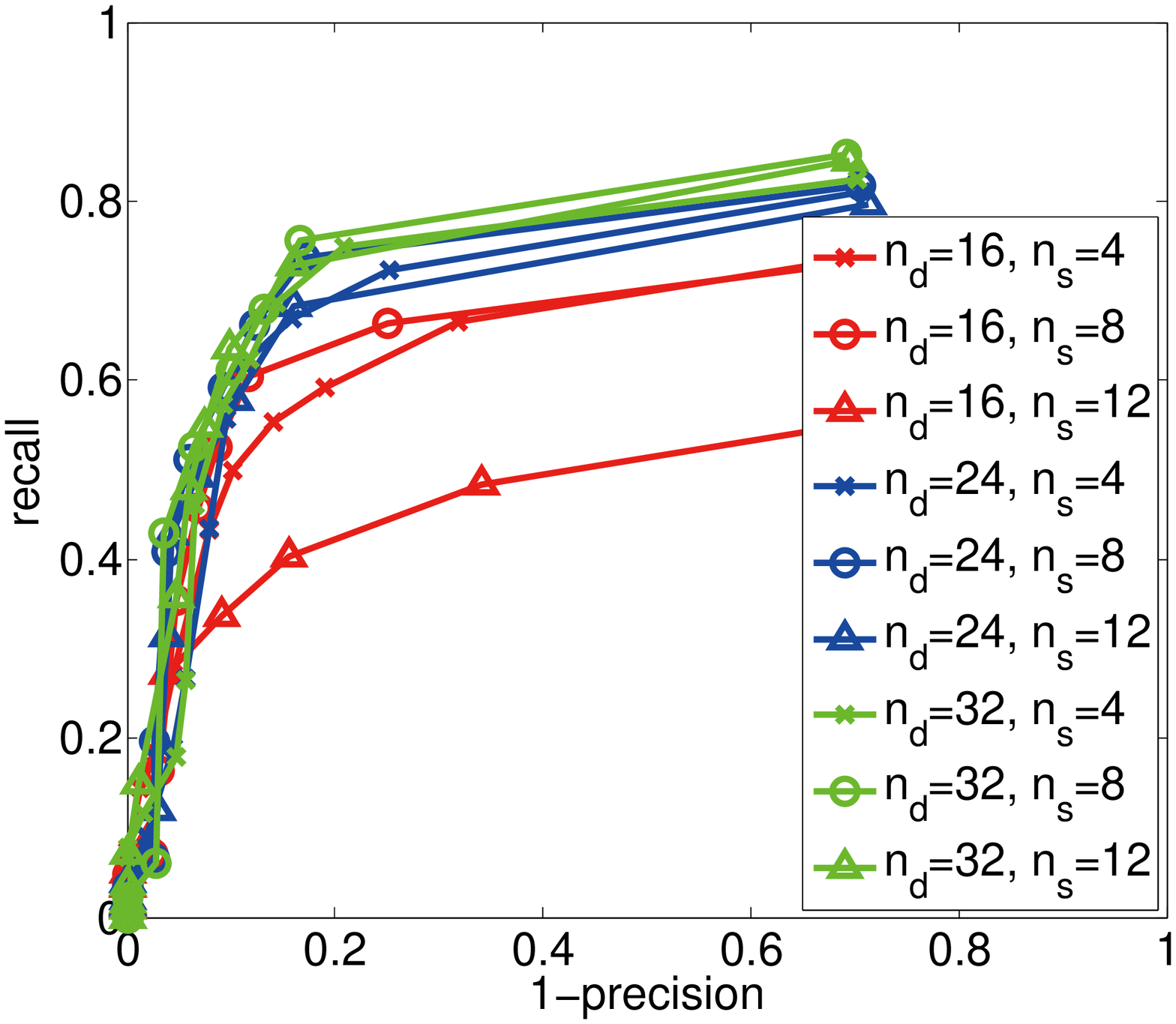}\label{subfig:parameters-naive}}
\subfigure[varying $n_l$ when $n_d=24$ and $n_s=8$]
{\includegraphics[width=0.48\textwidth]{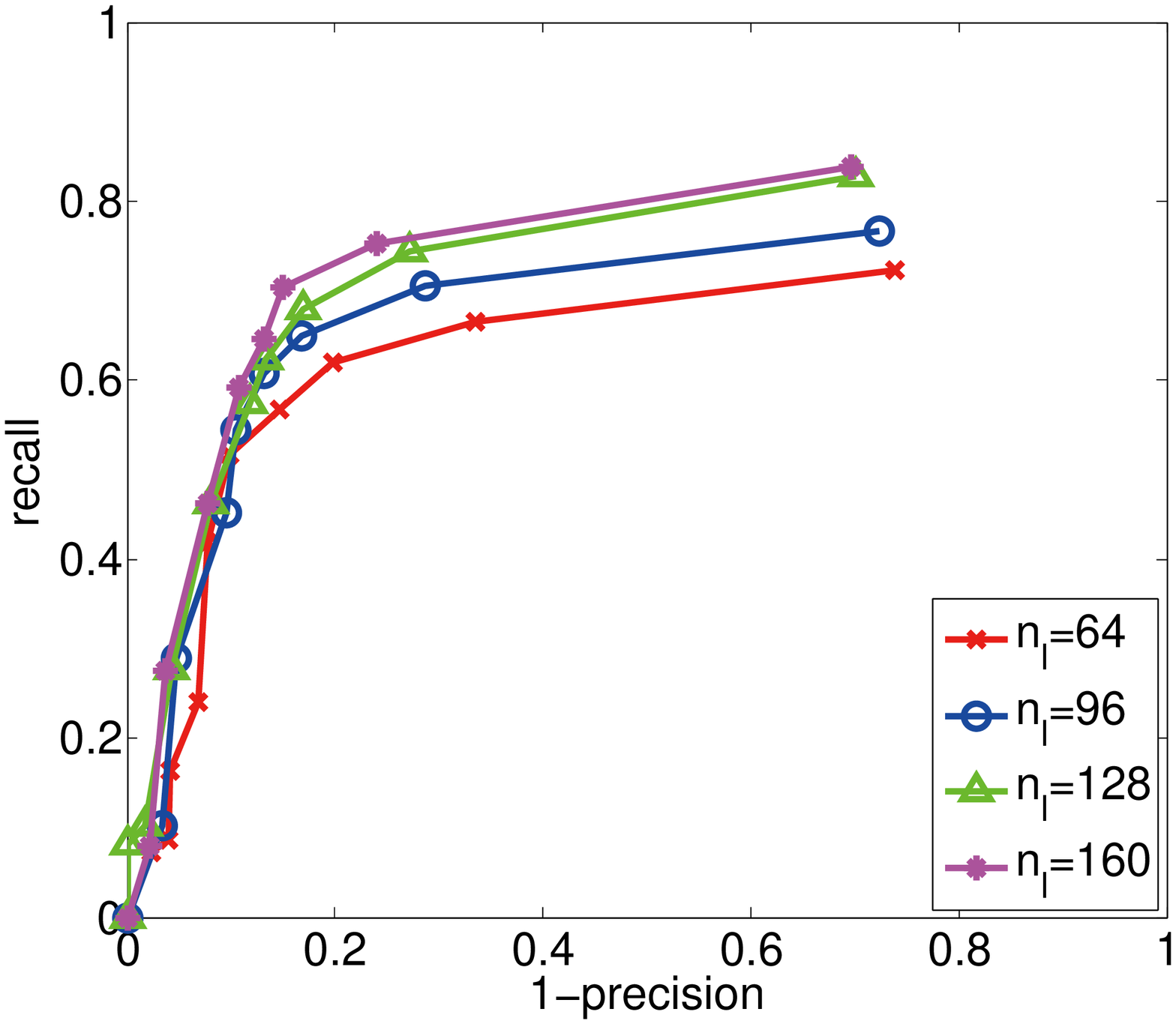}\label{subfig:parameters-fast}}
 \caption{Performance comparison of ASR descriptor on DoG keypoints under different parameter configurations by varying $n_d$, $n_s$ and  $n_l$.
\label{fig:parameters_2}}
\end{minipage}
\end{figure}

\subsection{Parameters Selection}
In addition to $T_o$ and $n_t$ for sampling affine transformations, our method has several other parameters listed in Table\ref{tb:parameters}.
We have investigated the effect of different parameter
settings on image pair of \emph{'trees 1-2'} in the Oxford
dataset~\cite{Oxford}. We simply tried several combinations
of these parameters and compared the matching performance among them.
The result is shown in Fig.~\ref{fig:parameters_2}.
Fig.~\ref{subfig:parameters-naive} is obtained by computing ASR-naive under different $n_d$ ($16$, $24$ and $32$) and $n_s$ ($4$, $8$ and $12$).
It is found that the configuration of $(n_d=32, n_s=8)$ obtains the best result.
For a trade off between the performance and descriptor dimension,
we choose $(n_d=24, n_s=8)$, leading to ASR with $24*(24+1)/2=300$ dimensions.
Under the choice of $(n_d=24, n_s=8)$,
we investigate the fast approximate algorithm by computing ASR-fast under different $n_l$ ($64$, $96$, $128$ and $160$).
Fig.~\ref{subfig:parameters-fast} shows that $n_l=160$ obtains the best result.
A typical setting of all parameters is given in Table~\ref{tb:parameters}
and kept unchanged in the subsequent experiments.

\begin{table}[htb]
\centering
\begin{tabular}{|c|c|c|}
\hline
parameter & description & typical value  \\
\hline
$n_p$    &  pattern number for dominant orientation estimation  & 60 \\
\hline
$n_l$    & number of orthogonal basis for approximating local patch & 160 \\
\hline
$s_l$    & size of local patch        & 21 \\
\hline
$n_d$    & dimension of the PCA-patch vector & 24 \\
\hline
$n_s$    & dimension of the subspace that PCA-patch vector set $\mathcal{D}$ lies on & 8 \\
\hline
\end{tabular}
\caption{Parameters in ASR descriptor and their typical settings.}
\label{tb:parameters}
\end{table}

\subsection{Evaluation on Oxford Dataset\label{sec:oxford}}
To show the superiority of our method, we conduct evaluations on this benchmark dataset based on the standard protocol~\cite{Mikolajczyk:2005:descriptor}, using the nearest neighbor distance ratio (NNDR) matching strategy. For comparison, the proposed method is compared with SIFT~\cite{LOWE:2004} and DAISY~\cite{winder2009picking} descriptors, which are the most popular ones representing the state-of-the-art. The results of other popular descriptors~(SURF, ORB, BRISK etc.) are not reported as they are inferior to that of DAISY.
In this experiment, keypoints are detected by DoG~\cite{LOWE:2004}
which is the most representative and widely used scale invariant detector.
Due to space limit, only the results of two image pairs (the $1^{st}$ vs. the $2^{nd}$ and the
$1^{st}$ vs. the $4^{th}$) for each image sequence are shown, which represent small and large image transformations respectively.

\begin{figure}[htb]
\centering
\subfigure[bikes 1-2]
{\includegraphics[width=0.24\textwidth,height=0.20\textwidth]{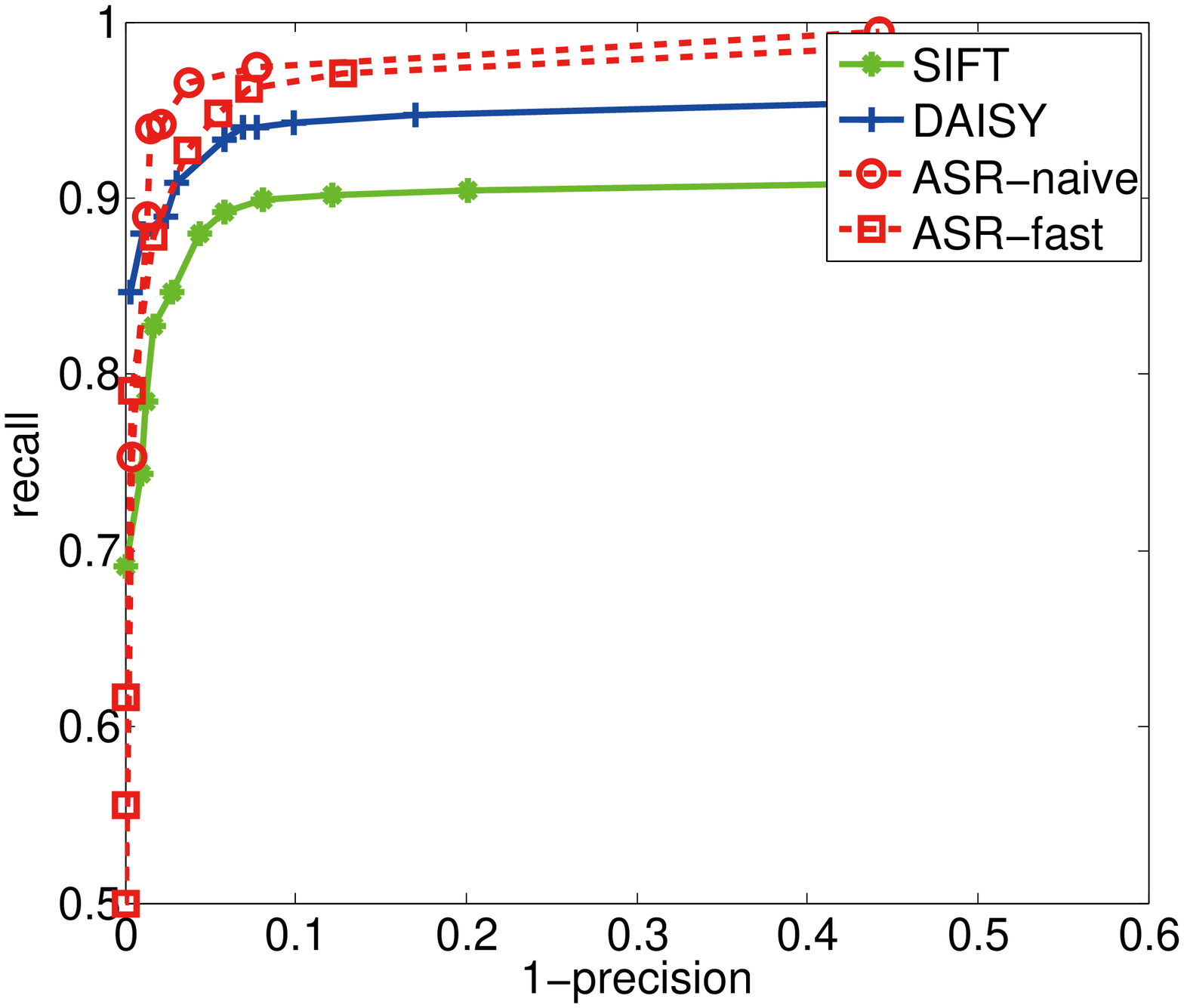}\label{subfig:dog_bikes1-2}}
\subfigure[bikes 1-4]
{\includegraphics[width=0.24\textwidth,height=0.20\textwidth]{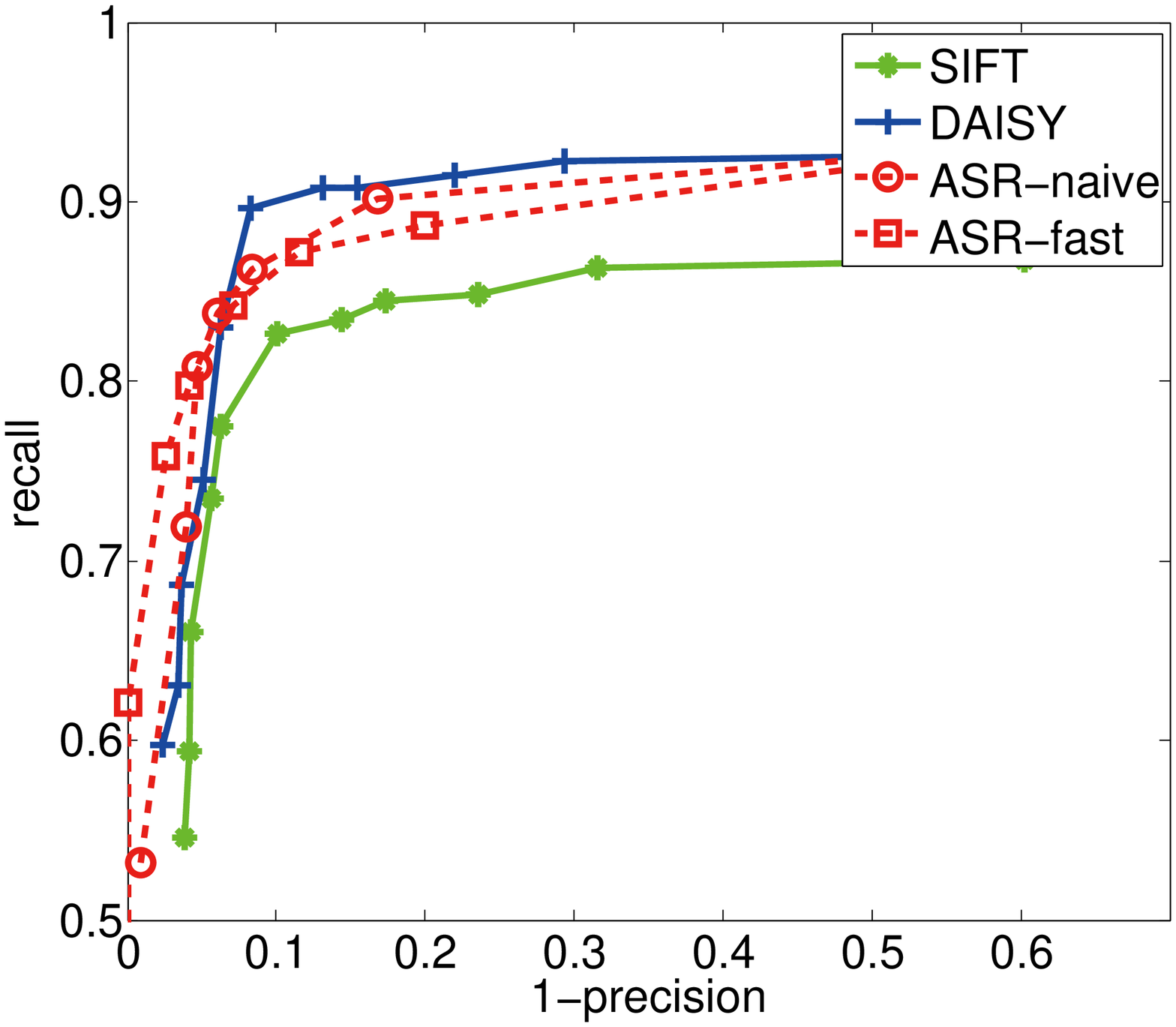}\label{subfig:dog_bikes1-4}}
\subfigure[boat 1-2]
{\includegraphics[width=0.24\textwidth,height=0.20\textwidth]{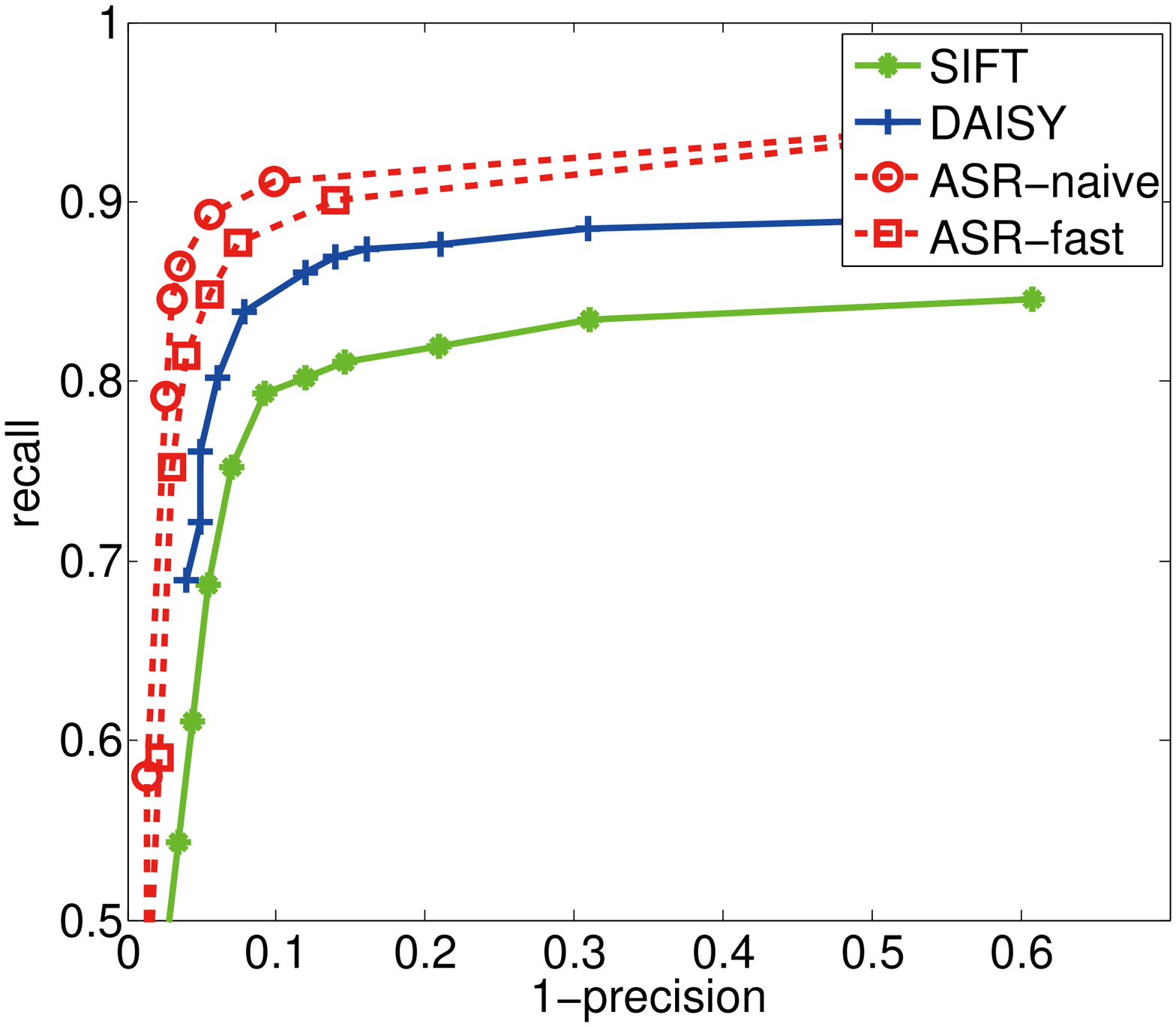}\label{subfig:dog_boat1-2}}
\subfigure[boat 1-4]
{\includegraphics[width=0.24\textwidth,height=0.20\textwidth]{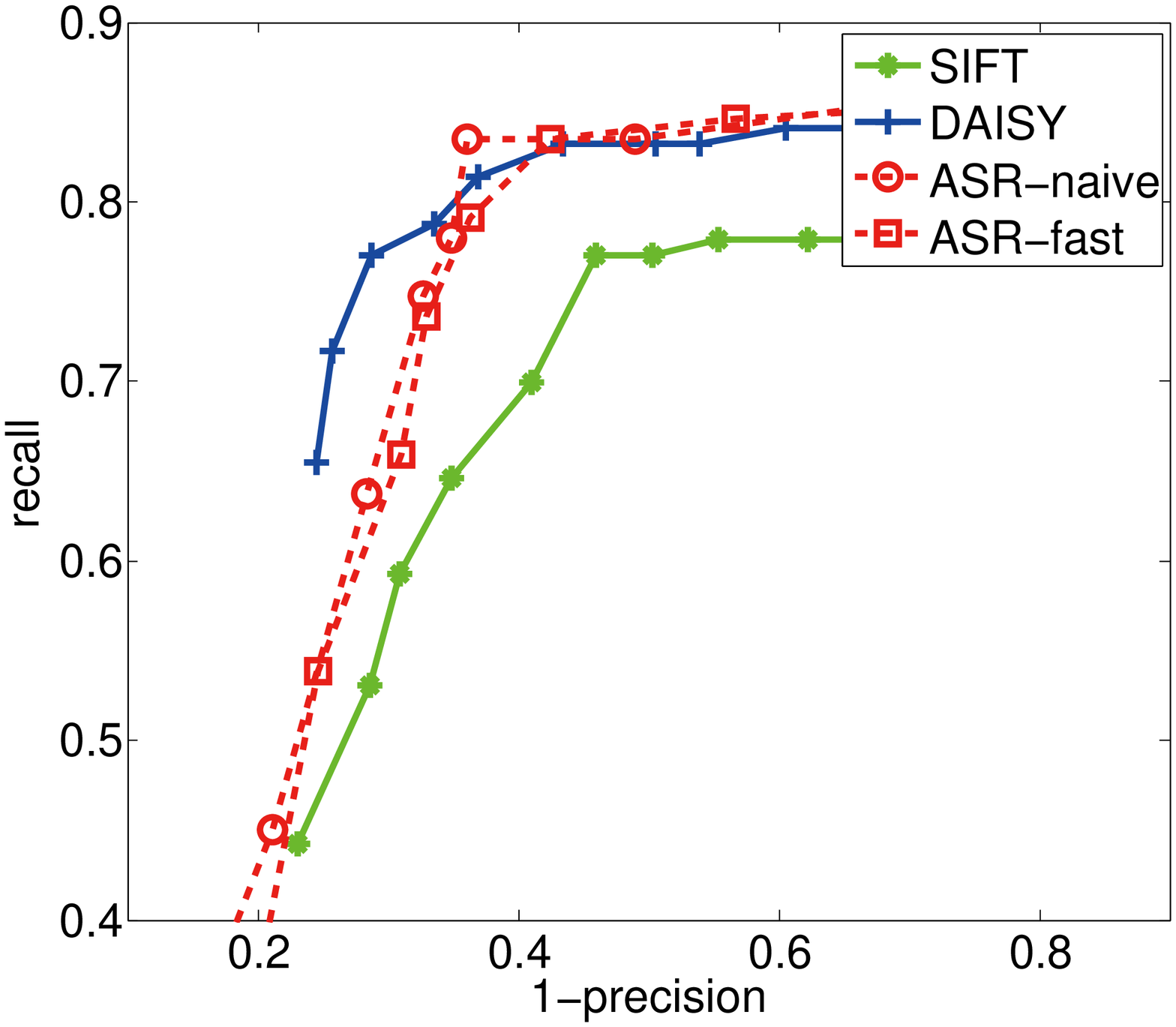}\label{subfig:dog_boat1-4}}
\subfigure[graf 1-2]
{\includegraphics[width=0.24\textwidth,height=0.20\textwidth]{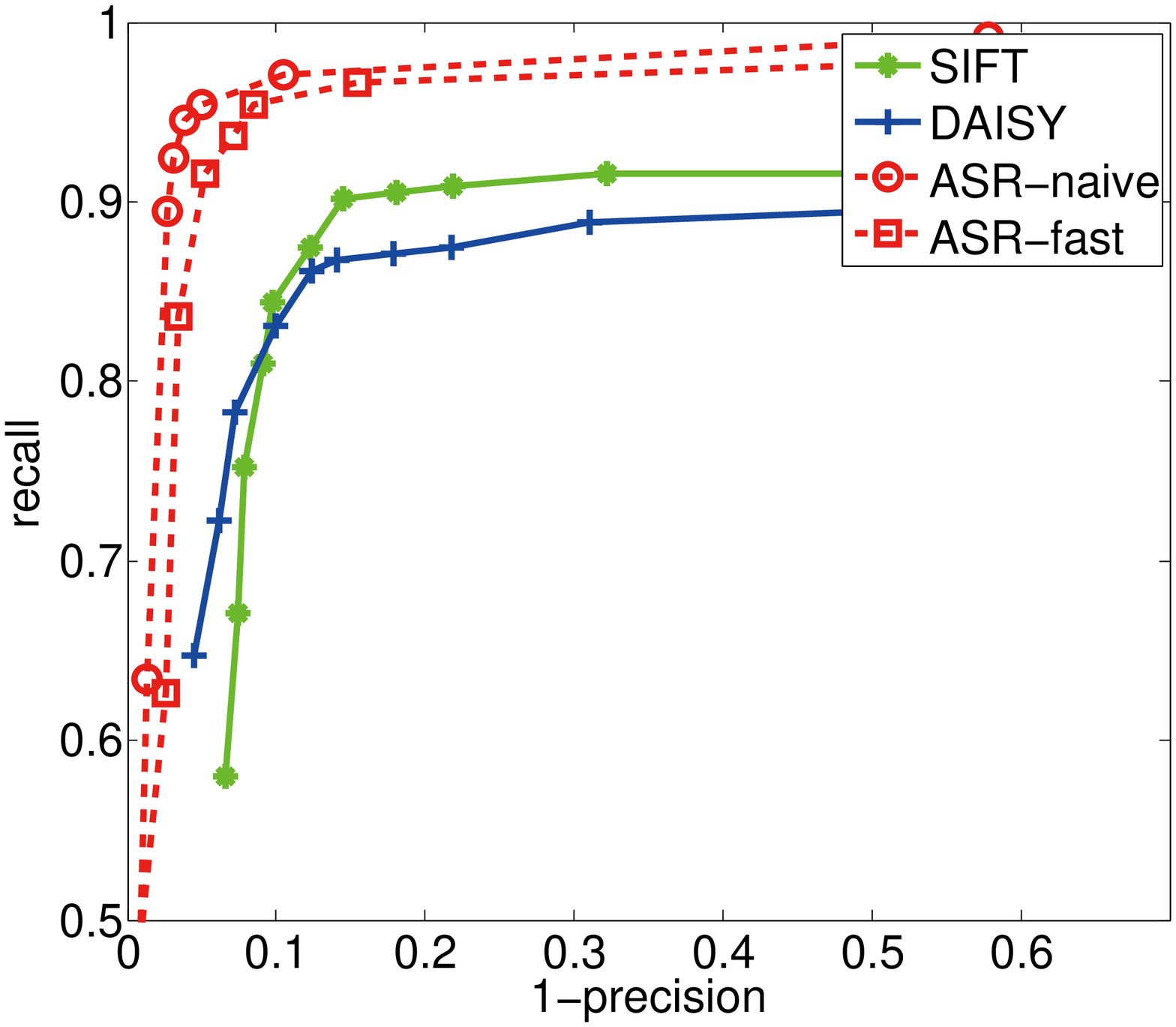}\label{subfig:dog_graf1-2}}
\subfigure[graf 1-4]
{\includegraphics[width=0.24\textwidth,height=0.20\textwidth]{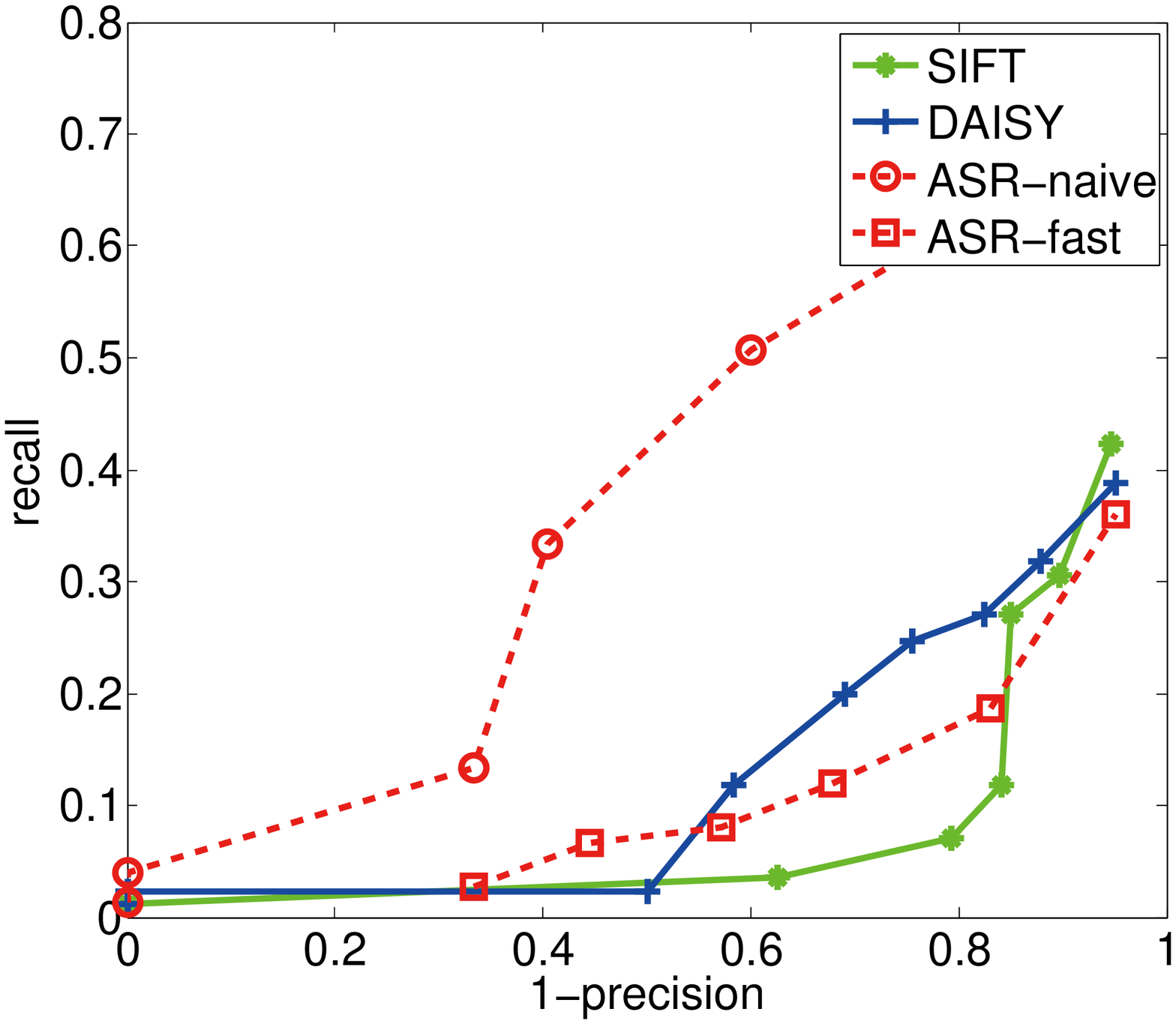}\label{subfig:dog_graf1-4}}
\subfigure[wall 1-2]
{\includegraphics[width=0.24\textwidth,height=0.20\textwidth]{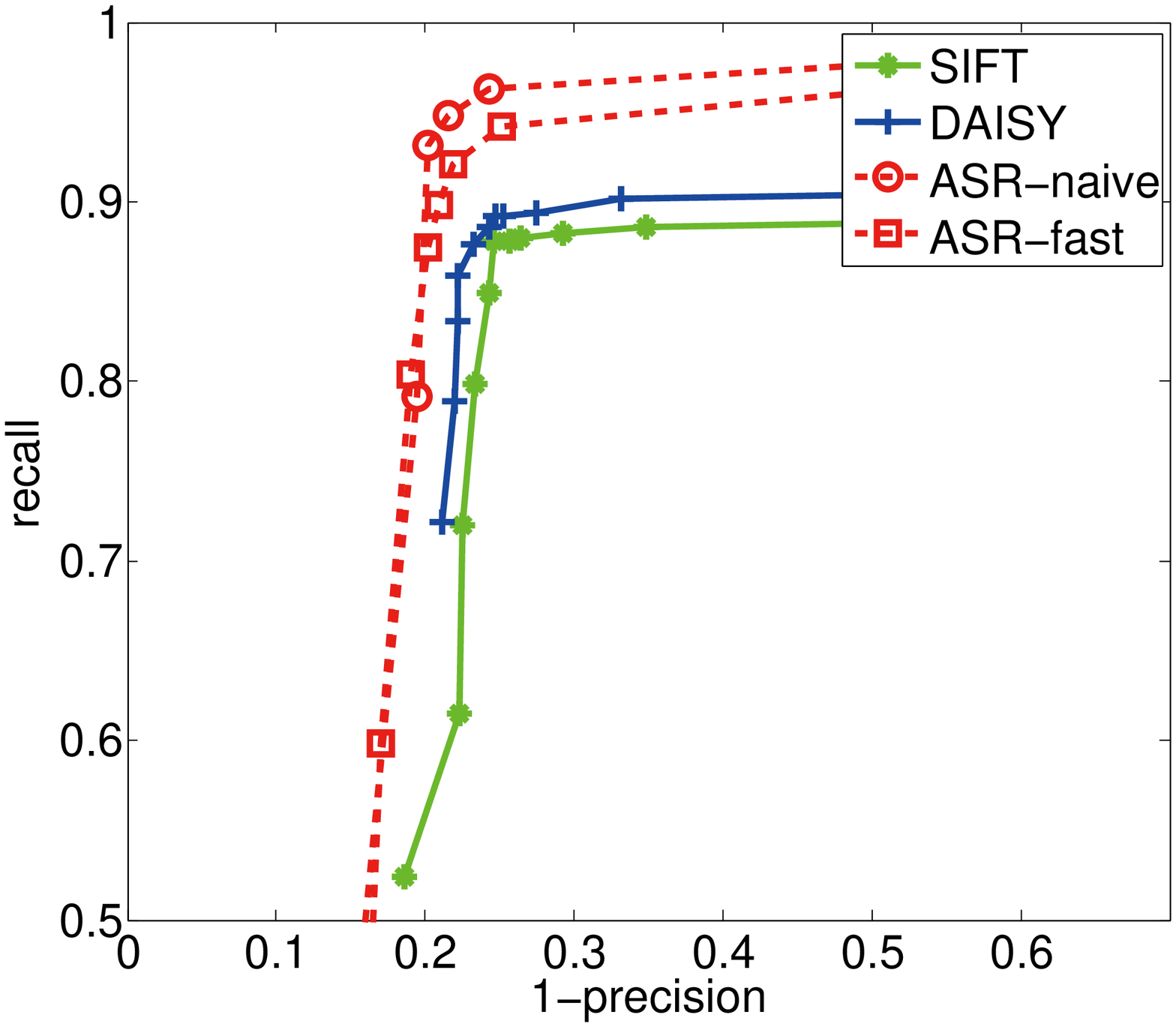}\label{subfig:dog_wall1-2}}
\subfigure[wall 1-4]
{\includegraphics[width=0.24\textwidth,height=0.20\textwidth]{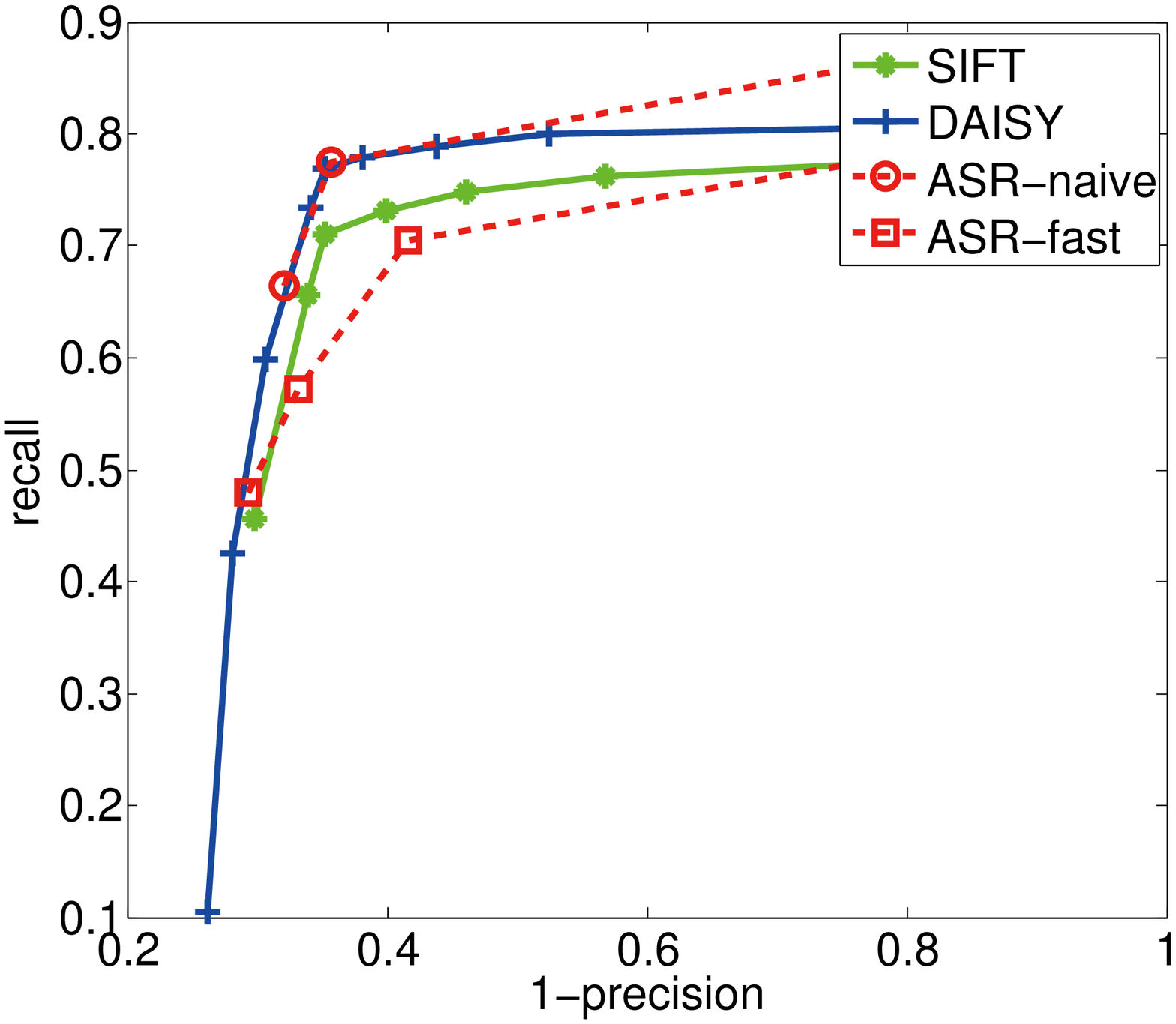}\label{subfig:dog_wall1-4}}
\subfigure[leuven 1-2]
{\includegraphics[width=0.24\textwidth,height=0.20\textwidth]{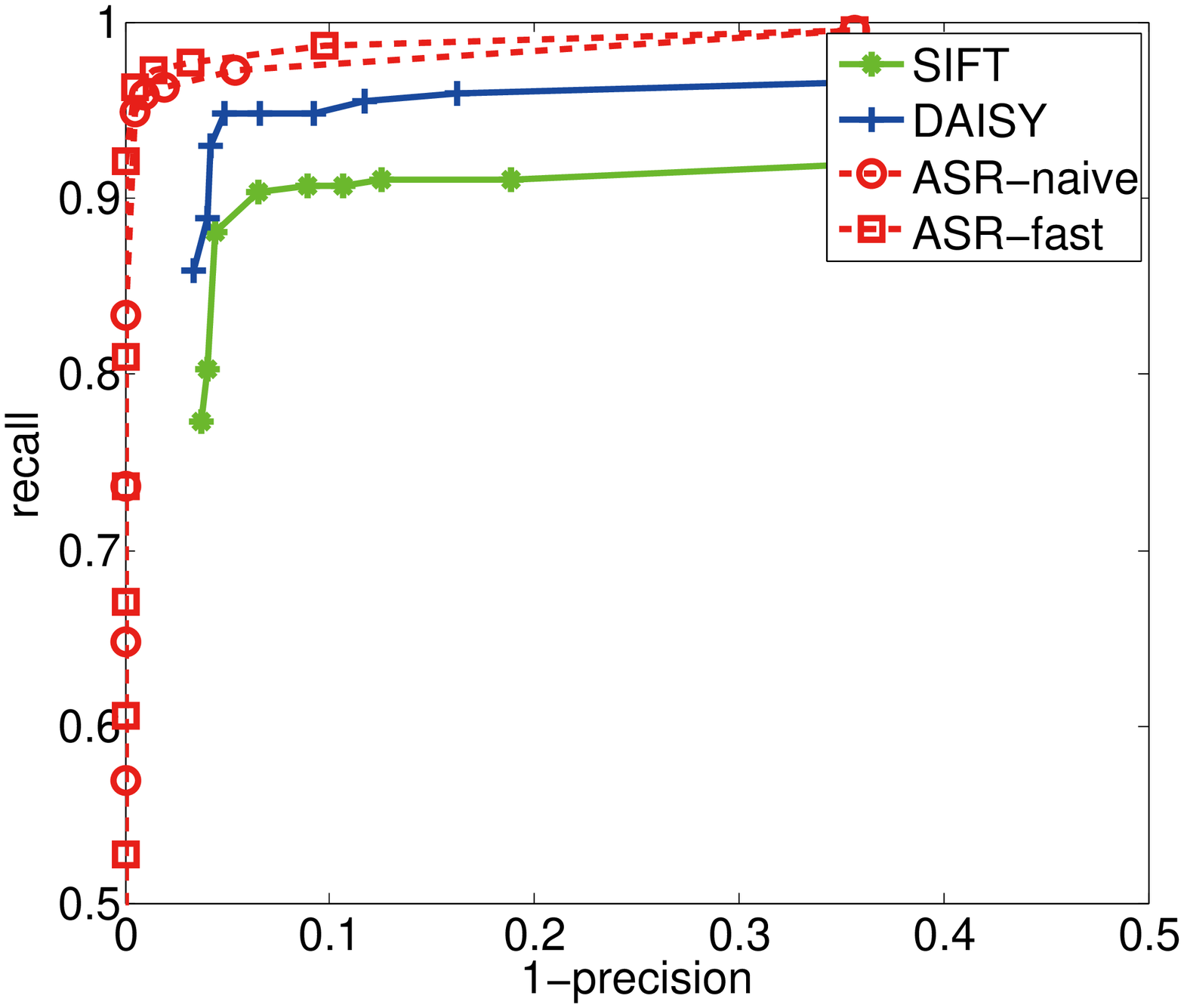}\label{subfig:dog_leuven1-2}}
\subfigure[leuven 1-4]
{\includegraphics[width=0.24\textwidth,height=0.20\textwidth]{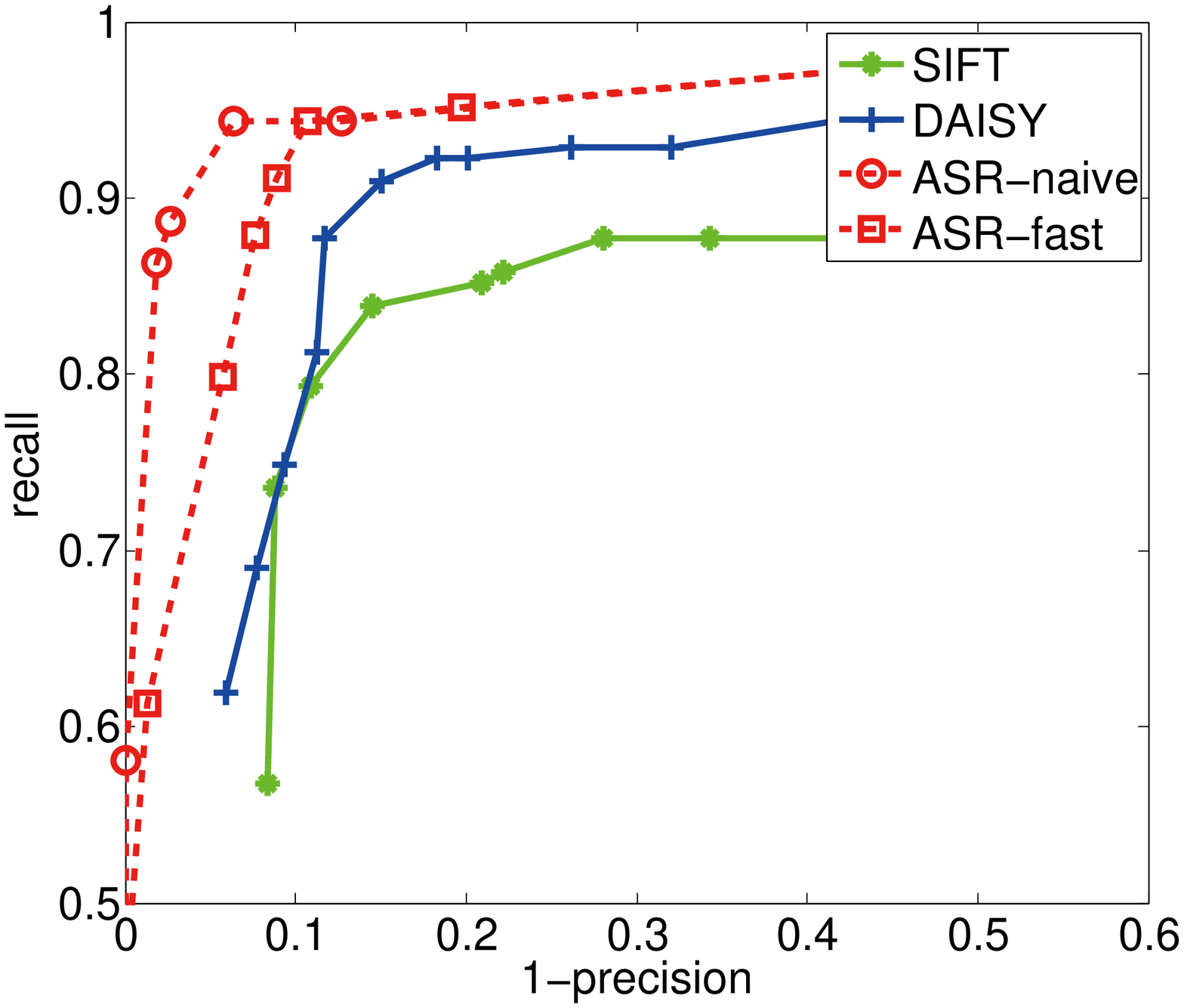}\label{subfig:dog_leuven1-4}}
\subfigure[ubc 1-2]
{\includegraphics[width=0.24\textwidth,height=0.20\textwidth]{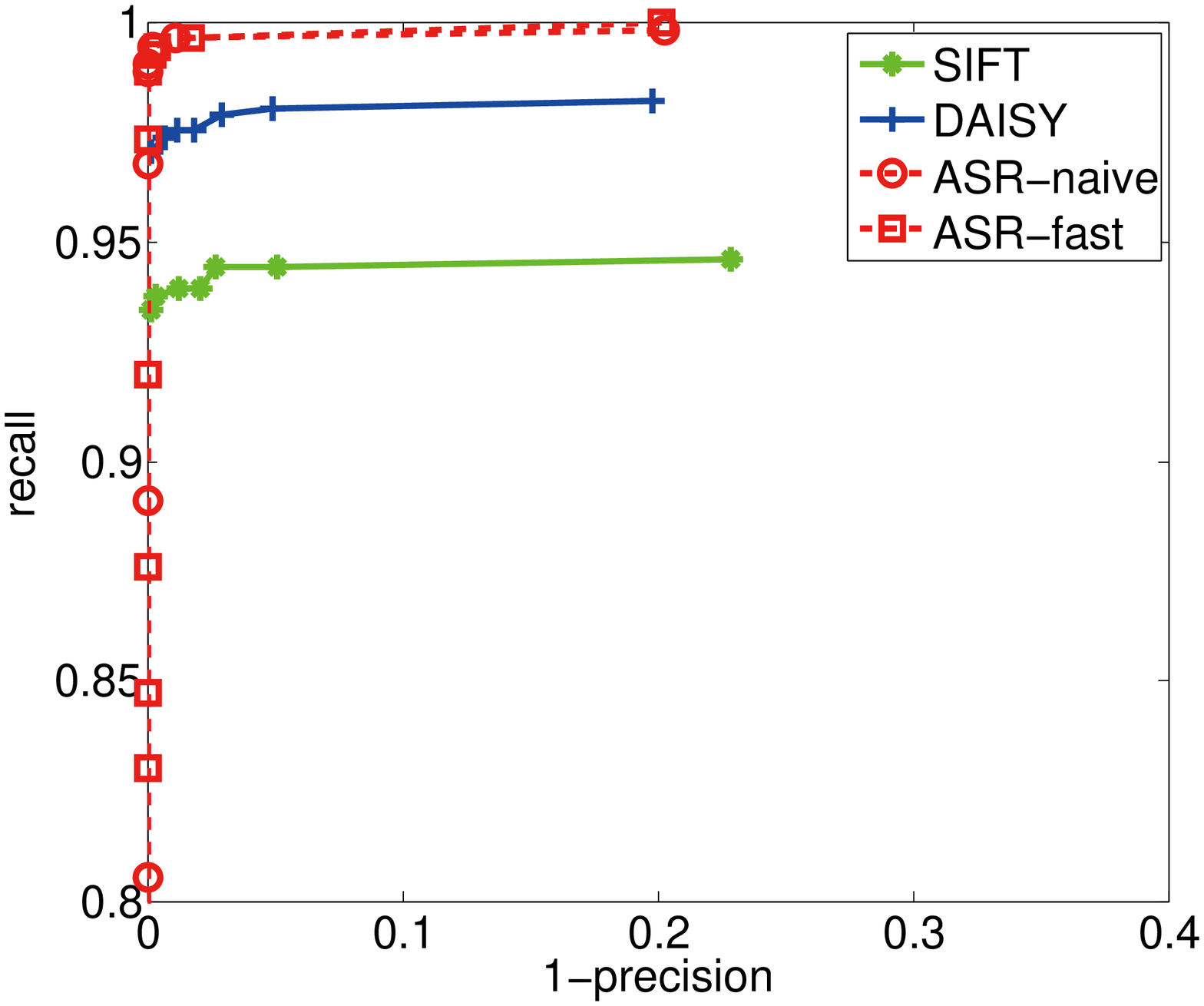}\label{subfig:dog_ubc1-2}}
\subfigure[ubc 1-4]
{\includegraphics[width=0.24\textwidth,height=0.20\textwidth]{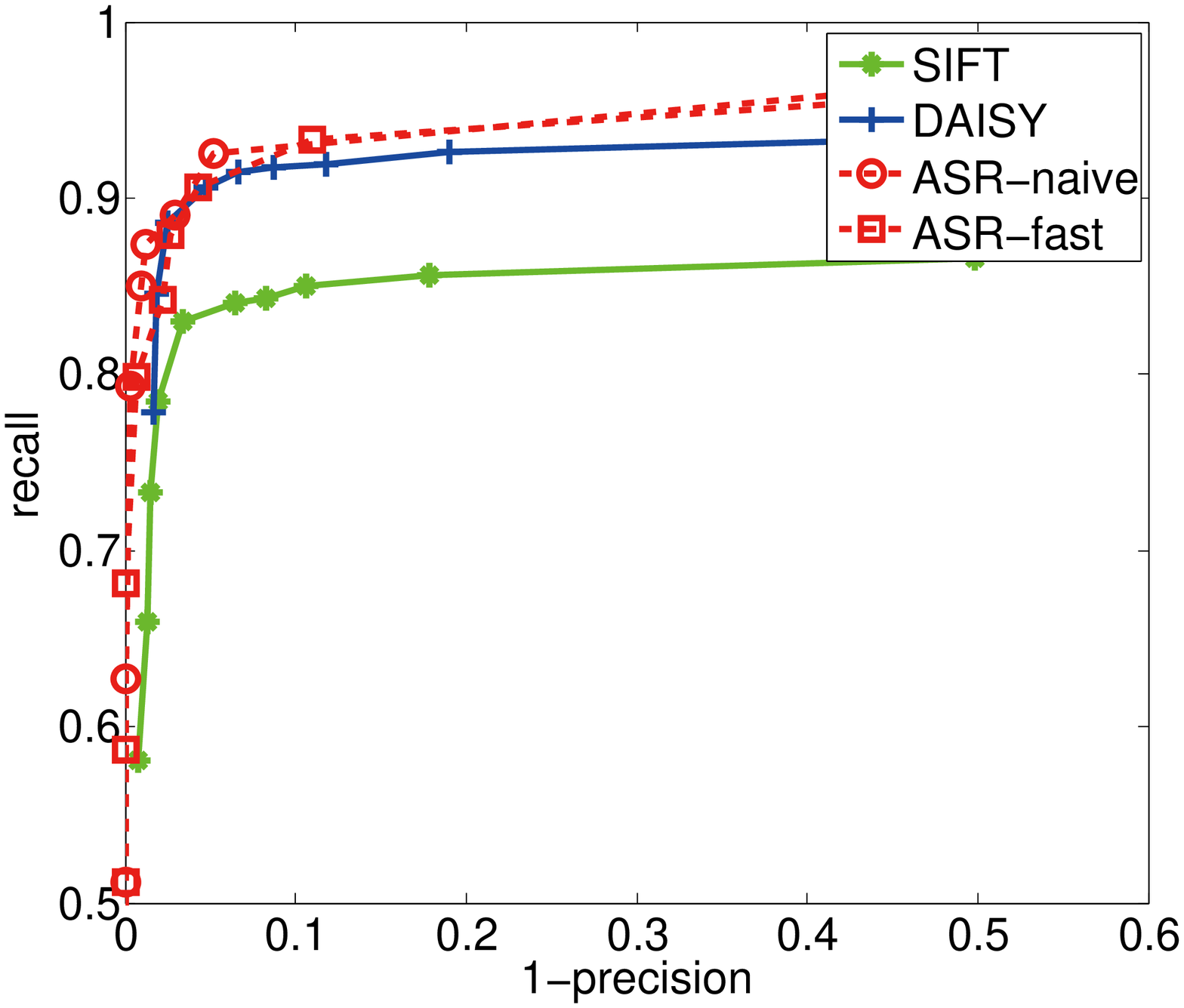}\label{subfig:dog_ubc1-4}}

\caption{Experimental results for different image transformations on DoG keypoints: (a)-(b) image blur,
(c)-(d) rotation and scale change, (e)-(h) viewpoint change, (i)-(j) illumination change and
(k)-(l) JPEG compression.}
\label{fig:dog_precision_recall}
\end{figure}

As shown in Fig.~\ref{fig:dog_precision_recall},
it is clear that ASR-fast performs comparable to ASR-naive in all cases except \emph{'graf 1-4'} (Fig.~\ref{subfig:dog_graf1-4}).
This demonstrates the fact that the proposed fast computation strategy in Eq. (\ref{eq:DA_final})
can well approximate the naive computation of PCA-patch vector set.
The performance degradation in \emph{'graf 1-4'} can be explained
by the difference in patch alignment. Since ASR-fast does not generate the warped patches directly,
it simply aligns the reference patch before computing the PCA-patch vector set.
This strategy could be unreliable under large image distortions since
all the PCA-patch vectors extracted under various affine transformations depend on
the orientation estimated on reference patch.
ASR-naive avoids this by computing the dominant orientation on each warped patch and aligning it separately. In other words, the inferior performance of ASR-fast is because that the PCA-patch vector (\ie, the intermediate representation) relies on robust orientation estimation, but \emph{does not} imply that ASR-fast is not suitable for viewpoint changes. Therefore, if we can use an inherent rotation invariant intermediate representation (such as the one in similar spirit to the intensity order based methods~\cite{Fan_CVPR11,Fan_PAMI12,Wang_ICCV11}), ASR-fast is expected to be as good as ASR-naive. We would leave this for our future work.

According to Fig.~\ref{fig:dog_precision_recall},
 both ASR-naive and ASR-fast are consistently
better than SIFT in all cases
and outperform DAISY in most cases.
The superior performance of the proposed method can be attributed to
the effective use of local information under various affine transformation.
For all cases of viewpoint changes especially in \emph{'graf 1-4'}, ASR-naive outperforms all competitors by a large margin,
which demonstrates its ability of dealing with affine distortions.

To further show ASR's ability in dealing with affine distortions without a dedicated affine detector, we use image pairs containing viewpoint changes to compare ASR with traditional methods, i.e., build local descriptor on top of affine invariant regions. In this experiment, Harris-Affine (HarAff) is used for interest region detection and SIFT/DAISY descriptors are constructed on these interest regions. For a fair comparison, ASR is build on top of Harris-Laplace (HarLap) detector since Harris-Affine regions are build up on Harris-Laplace regions by an additional affine adaptive procedure. Therefore, such a comparison ensures a fair evaluation for two types of affine invariant image matching methods, i.e., one based on affine invariant detectors, while the other based on affine robust descriptors.

\begin{figure}[htb]
\centering
\subfigure[graf 1-2]
{\includegraphics[width=0.24\textwidth,height=0.20\textwidth]{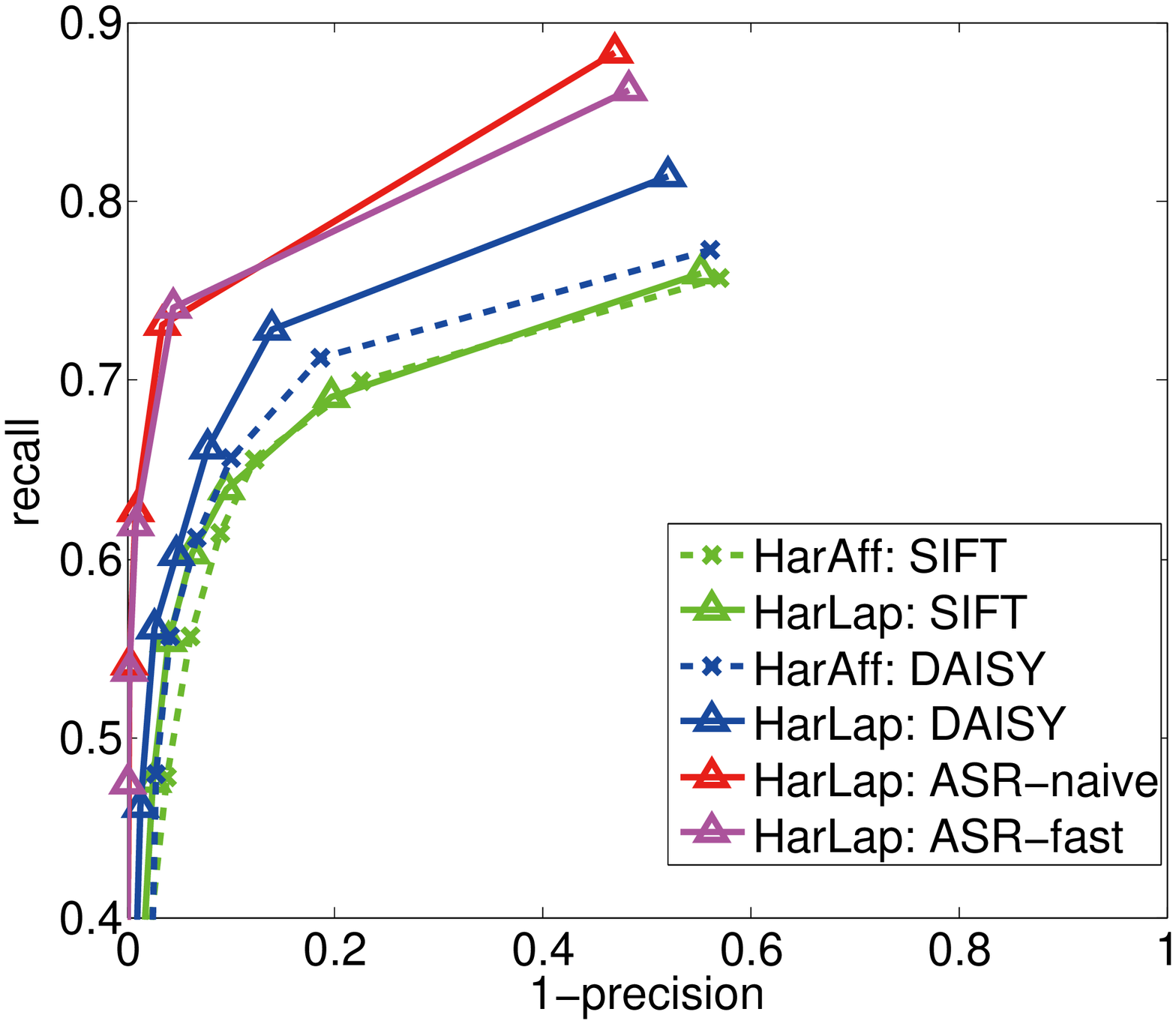}\label{subfig:affine_graf1-2}}
\subfigure[graf 1-4]
{\includegraphics[width=0.24\textwidth,height=0.20\textwidth]{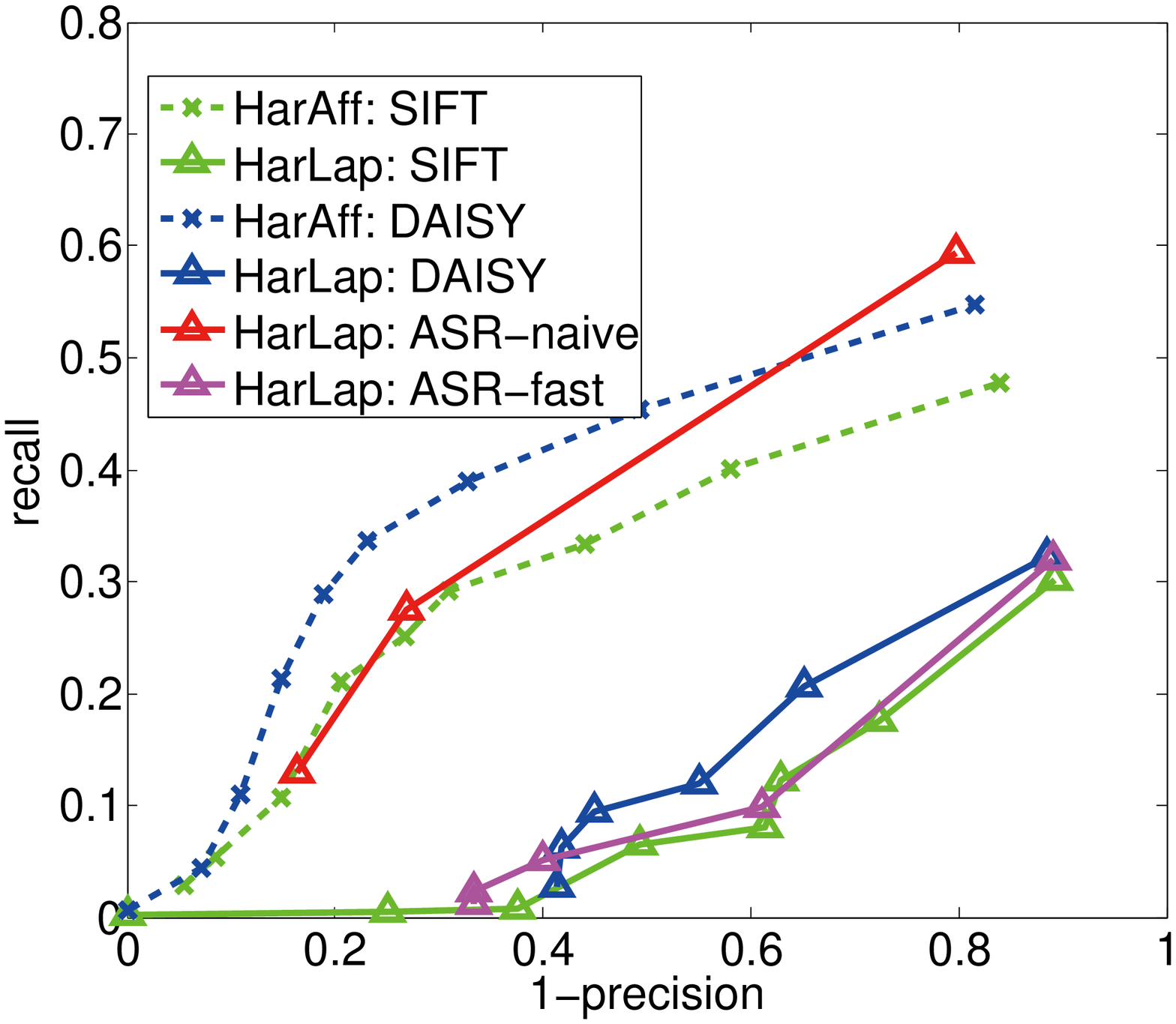}\label{subfig:affine_graf1-4}}
\subfigure[wall 1-2]
{\includegraphics[width=0.24\textwidth,height=0.20\textwidth]{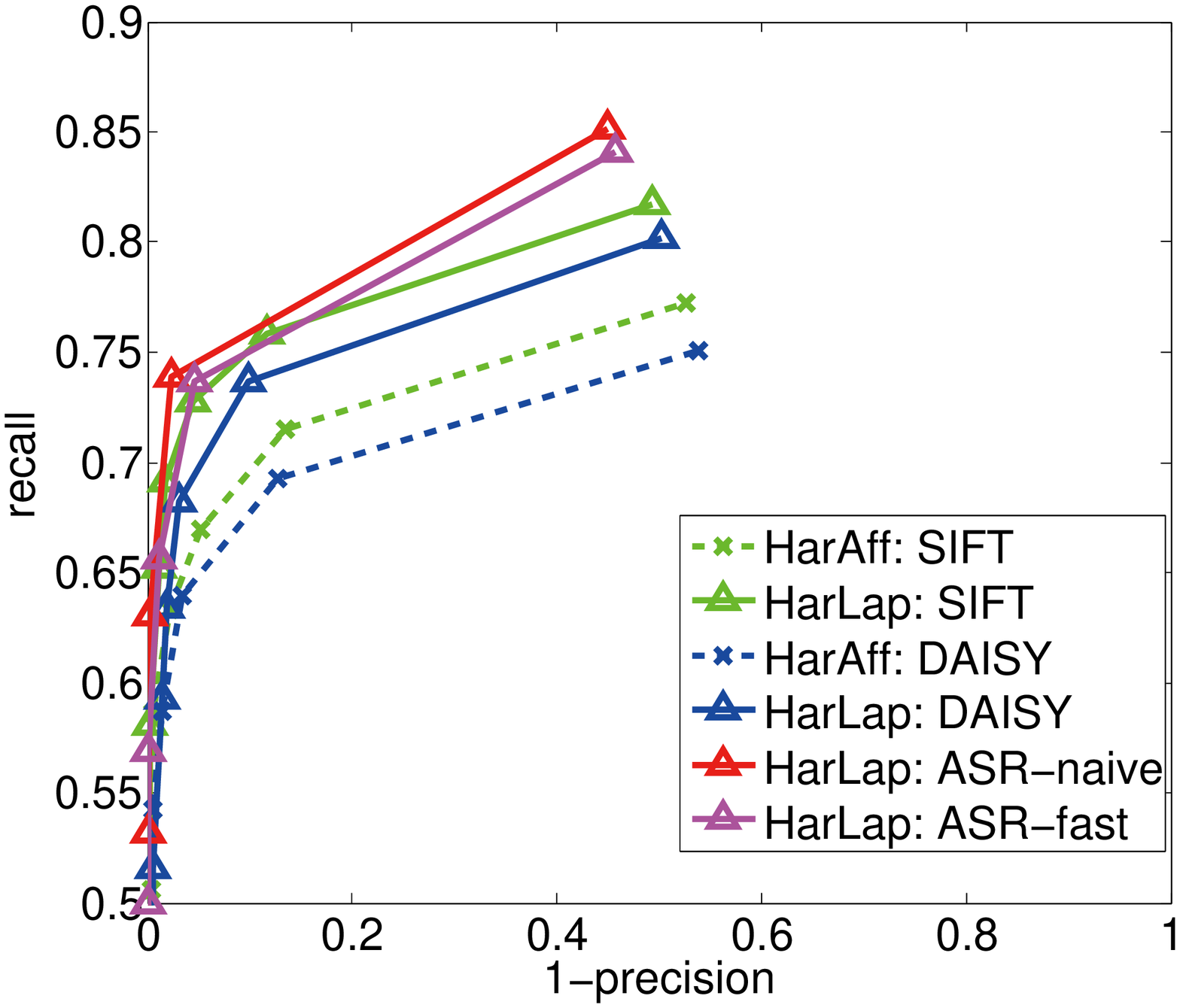}\label{subfig:affine_wall1-2}}
\subfigure[wall 1-4]
{\includegraphics[width=0.24\textwidth,height=0.20\textwidth]{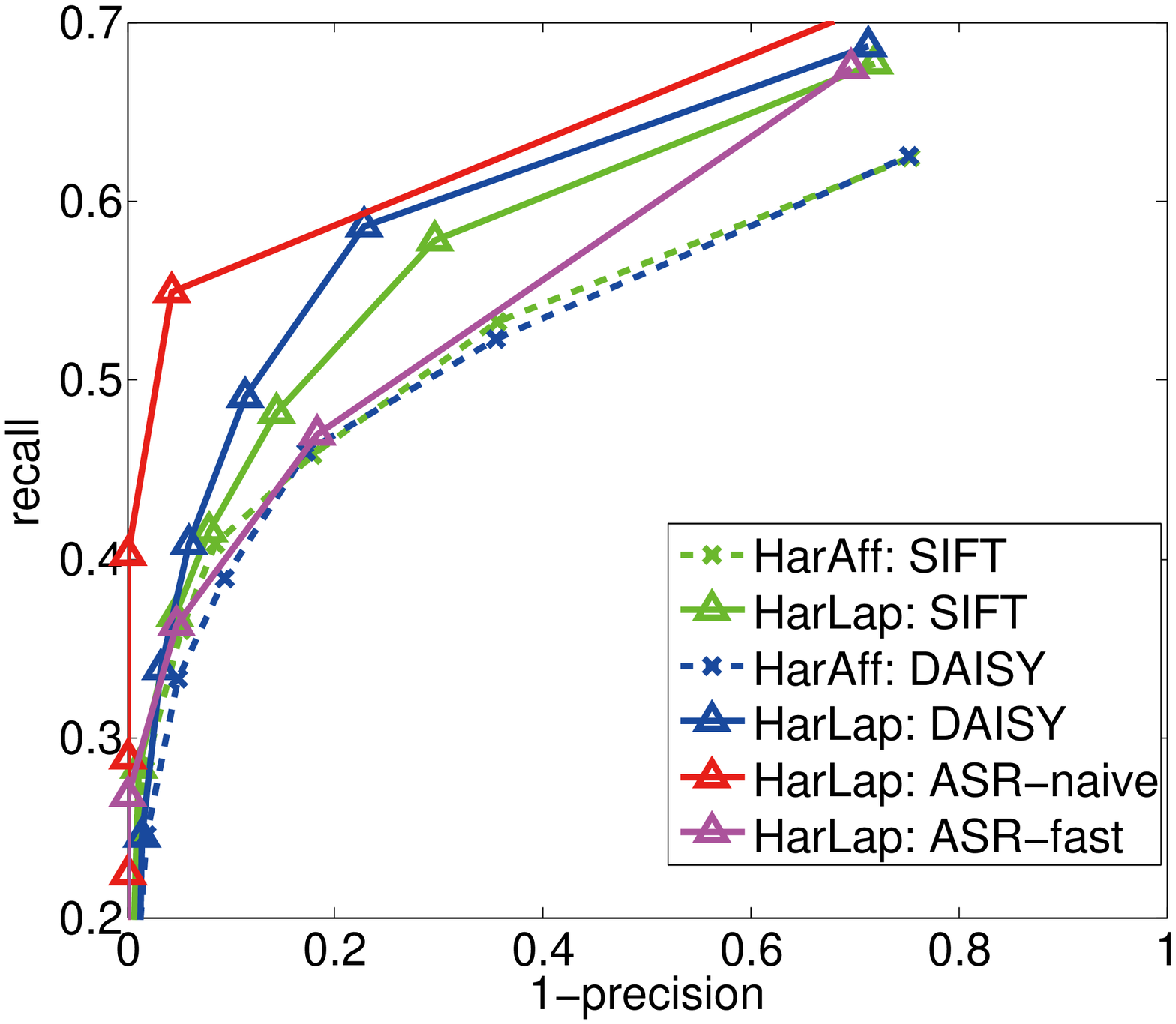}\label{subfig:affine_wall1-4}}
\caption{Experimental results on image sequences containing viewpoint changes.}
\label{fig:affine_precision_recall}
\end{figure}

The results are shown in Fig.~\ref{fig:affine_precision_recall}.
To show the affine adaptive procedure is necessary for dealing with affine distortions
if the used descriptor does not account for this aspect, the results of HarLap:SIFT and HarLap:DAISY are also supplied.
It is clear that HarAff:SIFT (HarAff:DAISY) is better than HarLap:SIFT (HarAff:DAISY).
By using the same detector, HarLap:ASR-naive significantly outperforms HarLap:SIFT and HarLap:DAISY.
It is also comparable to HarAff:DAISY and HarAff:SIFT in 'graf 1-4',  and even better than them in all other cases.
This demonstrate that by considering affine distortions in feature description stage, ASR is capable of matching images with viewpoint changes without a dedicated affine invariant detector. The failure of HarLap:ASR-fast is due to the unreliable orientation estimation as explained before.

Another excellent method to deal with affine invariant image matching problem is ASIFT. However, ASIFT can not be directly compared to the proposed method. This is because that ASIFT is an image matching framework while the propose method is a feature descriptor. Therefore, in order to give the reader a picture of how our method performs in image matching compared to ASIFT, we use ASR descriptor combined with DoG detector for image matching and the NNDR threshold is set to 0.8. The matching results are compared to those obtained by ASIFT when the matching threshold equals to 0.8. ASIFT is downloaded from the authors' website. The average matching precisions of all the image pairs in this dataset are 64.4\%, 80.8\% and 75.6\% for ASIFT, ASR-naive, and ASR-fast respectively. Accordingly, the average matching times of these methods are 382.2s, 14.5s and 8.3s when tested on the 'wall' sequence. We also note that the average matches are several hundreds when using ASR while they are one magnitude more when using ASIFT. Detailed matching results can be found in the supplemental material.

\subsection{Evaluation on 3D Object Dataset}
To obtain a more thoroughly study of dealing with affine distortions,
we have also evaluated our method on the 3D object dataset~\cite{Moreels:2007}, which has lots of images of 100 3D objects captured under various viewpoints. We use the same evaluation protocol as~\cite{Moreels:2007}.
The ROC curves are obtained by varying the threshold $T_{app}$ on the quality of the appearance match
, while the stability curves are obtained at fixed false alarm rate of $1.5*10^{-6}$.

As previous experimental setting, we use Harris-Laplace (HarLap) detector to produce scale invariant regions and then compute ASR descriptors for matching. For comparison, the corresponding Harris-Affine (HarAff) detector is used to produce affine invariant regions and SIFT/DAISY descriptors are computed based on them.

\begin{figure}[htb]
\centering
\subfigure[ROC curves]
{\includegraphics[width=0.24\textwidth,height=0.20\textwidth]{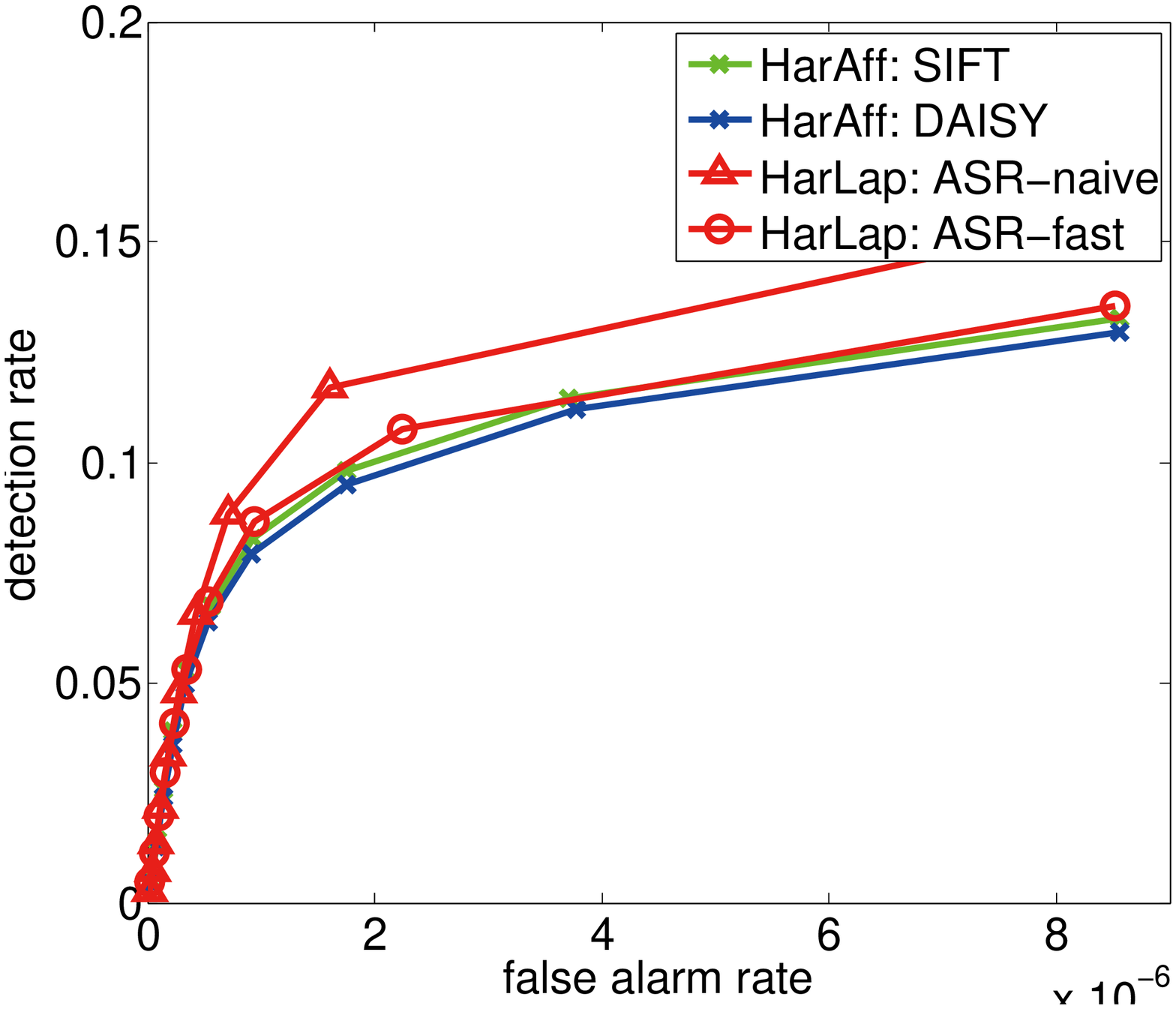}\label{subfig:3d roc har}}
\subfigure[Stability curves]
{\includegraphics[width=0.24\textwidth,height=0.20\textwidth]{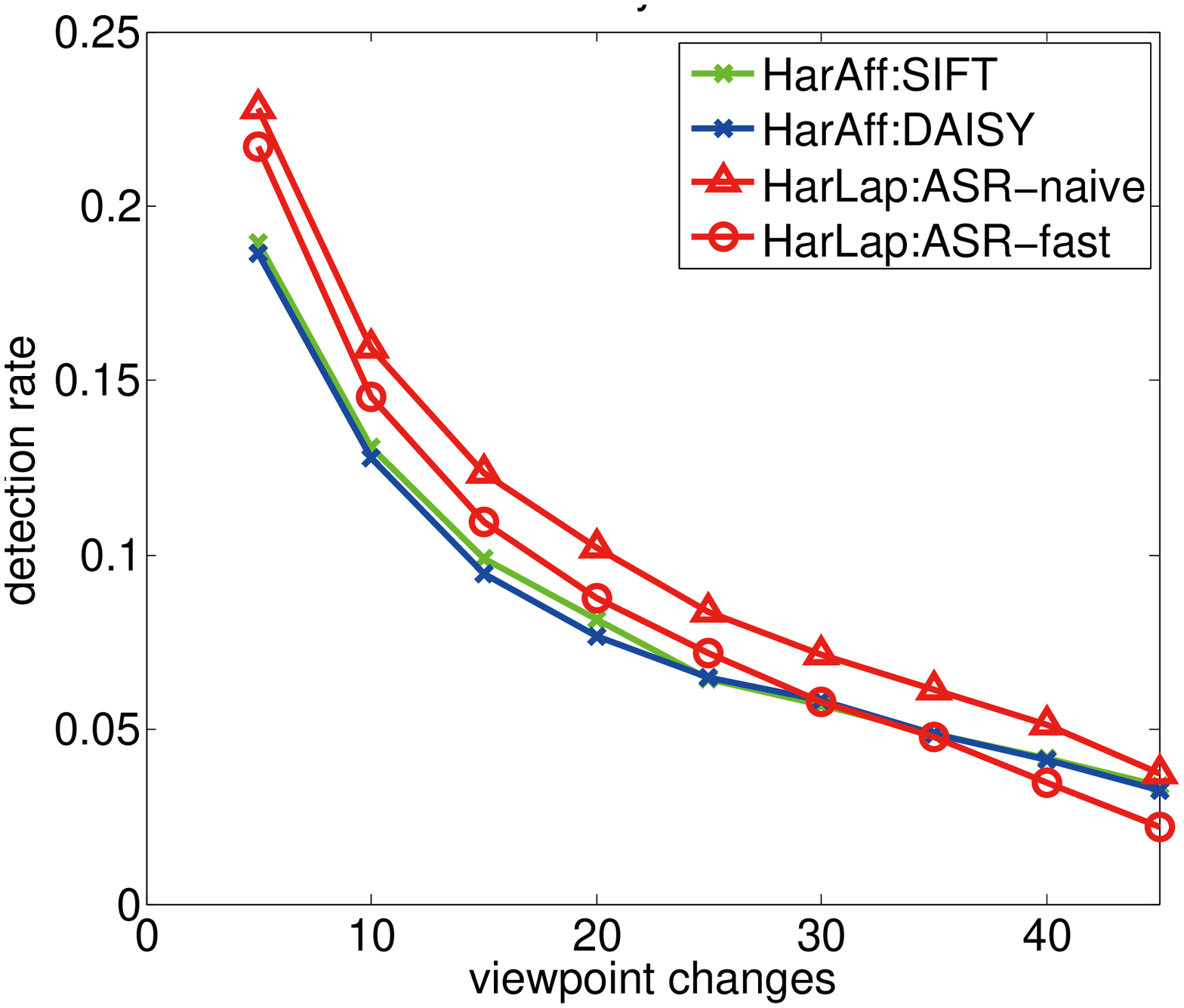}\label{subfig:3d stab har}}
\caption{Performance of different methods for 3D Object Dataset.
\label{fig:3d roc and stab}}
\end{figure}

Fig.~\ref{fig:3d roc and stab} shows the results averaged on all objects in the dataset when the viewing angle is varied from $5^o$ to $45^o$.
It can be observed that HarLap:ASR-naive performs best, and  HarLap:ASR-fast is comparable to  HarAff:SIFT and HarAff:DAISY.
This further demonstrates that the subspace representation of PCA-patch vectors extracted under various affine transformations is capable of dealing with affine distortion.

\subsection{Timing Result}
In this section, we conduct time test on a desktop with an Intel Core2
Quad 2.83GHz CPU. We first test the time cost for each components of ASR, and
the detailed results are given in Table~\ref{tb:timing}. It can be found that most of construction time in ASR is spent on patch warping. It is worthy to note that by using the fast approximate algorithm, ASR-fast does not compute the warped patch directly
and so largely reduce its time by about $75\%$. For comparison, we also report the time costs for SIFT and DAISY.
Note that these timing results are averaged over $100$ runs, each of which computes about 1000 descriptors on image 'wall 1'.
It is clear that ASR-fast is faster than  SIFT and DAISY,
while ASR-naive is slower than SIFT but still comparable to DAISY.

\begin{table}[htb]
\begin{center}
\begin{tabular}{|c|c|c|c|c|c|}
\hline
                                  & ASR-naive    &    ASR-fast    & SIFT  & DAISY  \\
\hline
patch warping[ms]                 & $2.98$       & $0.00$  &   -    &  -    \\\hline
patch representation[ms]          & $0.71$       & $0.64$  &     -    & -  \\\hline
subspace representation[ms]       & $0.49$       & $0.49$  &       -   &    -     \\\hline
total time[ms]                    & $4.18$       & $1.13$  &    2.09 &  3.8   \\\hline
\end{tabular}
\end{center}
\caption{Timing costs for constructing different descriptors.}
\label{tb:timing}
\end{table}

\section{Conclusion\label{sec:conclusion}}
In this paper, we have proposed the Affine Subspace Representation (ASR) descriptor.
The novelty lies in three aspects:
$1)$ dealing with affine distortion by integrating local information under multiple views,
which avoids the inaccurate affine shape estimation,
$2)$ a fast approximate algorithm for efficiently computing the PCA-patch vector of each warped patch,
and $3)$ the subspace representation of PCA-patch vectors extracted under various affine transformations of the same keypoint.

Different from existing methods, ASR effectively exploits the local information of a keypoint
by integrating the PCA-patch vectors of all warped patches.
The use of multiple views' information makes it is capable of dealing
with affine distortions to a certain degree while maintaining high distinctiveness.
What is more, to speedup the computation,
a fast approximate algorithm is proposed at a little cost of performance degradation.
Extensive experimental evaluations have demonstrated the effectiveness of the proposed method.

\section{Acknowledgment}
This work is supported by the National Nature Science Foundation of China~(No.91120012, 61203277, 61272394) and the Beijing Nature Science Foundation~(No.4142057).

\bibliographystyle{splncs03}
\bibliography{egbib}

\begin{thebibliography}{10}
\providecommand{\url}[1]{\texttt{#1}}
\providecommand{\urlprefix}{URL }

\bibitem{Oxford}
http://www.robots.ox.ac.uk/~vgg/research/affine/

\bibitem{basri2011}
Basri, R., Hassner, T., Zelnik-Manor, L.: Approximate nearest subspace search.
  PAMI  33(2),  266--278 (2011)

\bibitem{baumberg2000}
Baumberg, A.: Reliable feature matching across widely separated views. In:
  Proc. of CVPR. vol.~1, pp. 774--781. IEEE (2000)

\bibitem{BAY:2006}
Bay, H., Tuytelaars, T., Van~Gool, L.: Surf: Speeded up robust features. In:
  Proc. of ECCV. pp. 404--417 (2006)

\bibitem{edelman1998}
Edelman, A., Arias, T.A., Smith, S.T.: The geometry of algorithms with
  orthogonality constraints. SIAM journal on Matrix Analysis and Applications
  20(2),  303--353 (1998)

\bibitem{Fan_CVPR11}
Fan, B., Wu, F., Hu, Z.: Aggregating gradient distributions into intensity
  orders: A novel local image descriptor. In: Proc. of CVPR. pp. 2377--2384
  (2011)

\bibitem{Fan_PAMI12}
Fan, B., Wu, F., Hu, Z.: Rotationally invariant descriptors using intensity
  order pooling. PAMI  34(10),  2031--2045 (2012)

\bibitem{Hassner2012}
Hassner, T., Mayzels, V., Zelnik-Manor, L.: On sifts and their scales. In:
  Proc. of CVPR (2012)

\bibitem{Hinterstoisser2011}
Hinterstoisser, S., Lepetit, V., Benhimane, S., Fua, P., Navab, N.: Learning
  real-time perspective patch rectification. IJCV pp. 1--24 (2011)

\bibitem{Leutenegger2011}
Leutenegger, S., Chli, M., Siegwart, R.: Brisk: Binary robust invariant
  scalable keypoints. In: Proc. of ICCV. pp. 2548--2555 (2011)

\bibitem{LINDEBERG:1998}
Lindeberg, T.: Feature detection with automatic scale selection. IJCV  30(2),
  79--116 (1998)

\bibitem{lindeberg1997shape}
Lindeberg, T., G{\aa}rding, J.: Shape-adapted smoothing in estimation of 3-d
  shape cues from affine deformations of local 2-d brightness structure. Image
  and Vision Computing  15(6),  415--434 (1997)

\bibitem{LOWE:2004}
Lowe, D.: Distinctive image features from scale-invariant keypoints. IJCV
  60(2),  91--110 (2004)

\bibitem{MATAS:2002}
Matas, J., Chum, O., Urban, M., Stereo, T.: Robust wide baseline stereo from
  maximally stable extremal regions. Proc. of BMVC pp. 414--431 (2002)

\bibitem{MIKOLAJCZYK:2005}
Mikolajczyk, K., Tuytelaars, T., Schmid, C., Zisserman, A., Matas, J.,
  Schaffalitzky, F., Kadir, T., Gool, L.: A comparison of affine region
  detectors. IJCV  65(1),  43--72 (2005)

\bibitem{Mikolajczyk04}
Mikolajczyk, K., Schmid, C.: Scale \& affine invariant interest point
  detectors. IJCV  60,  63--86 (2004)

\bibitem{Mikolajczyk:2005:descriptor}
Mikolajczyk, K., Schmid, C.: A performance evaluation of local descriptors.
  PAMI  27(10),  1615--1630 (2005)

\bibitem{Moreels:2007}
Moreels, P., Perona, P.: Evaluation of features detectors and descriptors based
  on 3d objects. International Journal of Computer Vision  73(3),  263--284
  (2007)

\bibitem{morel2009}
Morel, J.M., Yu, G.: Asift: A new framework for fully affine invariant image
  comparison. SIAM Journal on Imaging Sciences pp. 438--469 (2009)

\bibitem{ozuysal2010fast}
Ozuysal, M., Calonder, M., Lepetit, V., Fua, P.: Fast keypoint recognition
  using random ferns. PAMI  32(3),  448--461 (2010)

\bibitem{Tuytelaars04}
Tuytelaars, T., Van~Gool, L.: Matching widely separated views based on affine
  invariant regions. IJCV  59,  61--85 (2004)

\bibitem{Wang_ICCV11}
Wang, Z., Fan, B., Wu, F.: Local intensity order pattern for feature
  description. In: Proc. of ICCV. pp. 603--610 (2011)

\bibitem{winder2009picking}
Winder, S., Hua, G., Brown, M.: Picking the best daisy. In: Proc. of CVPR. pp.
  178--185 (2009)

\bibitem{YAN2004}
Yan, K., Sukthankar, R.: Pca-sift: A more distinctive representation for local
  image descriptors. In: Proc. of CVPR. pp. 506--513 (2004)

\end{thebibliography}
\end{document}